\pdfoutput=1

\documentclass[11pt]{article}

\usepackage[final]{acl}
\usepackage{soul}
\usepackage{times}
\usepackage{latexsym}
\usepackage{enumitem}
\usepackage[T1]{fontenc}

\usepackage[utf8]{inputenc}

\usepackage{microtype}

\usepackage{inconsolata}

\usepackage{graphicx}
\usepackage{multirow}
\usepackage{amsmath,amssymb,amsthm}

\usepackage{hyperref}       
\usepackage{url}            
\usepackage{colortbl}
\usepackage{subfigure}
\usepackage{array}
\usepackage{tabularx}

\usepackage{booktabs}
\usepackage{xspace}
\usepackage{multirow}
\usepackage{makecell}
\usepackage{algorithm}
\usepackage{algpseudocode}
\newcommand{\mymethod}{Data Whisperer}
\definecolor{mycolor}{RGB}{218, 242, 251}

\usepackage[utf8]{inputenc}
\usepackage{adjustbox}

\usepackage{microtype}

\usepackage{inconsolata}

\usepackage{graphicx}

\title{\mymethod{}: Efficient Data Selection for Task-Specific LLM Fine-Tuning via Few-Shot In-Context Learning}

\author{
    Shaobo Wang$^{1,2}$\thanks{Project Head (\texttt{\href{mailto:shaobowang1009@sjtu.edu.cn}{shaobowang1009@sjtu.edu.cn}}).} \quad Xiangqi Jin$^{2\dagger}$ \quad  Ziming Wang$^{2,3}$\thanks{Xiangqi Jin and Ziming Wang are equally contributed as the second author.}  \quad Jize Wang$^1$ \quad Jiajun Zhang$^2$ \\ 
    \textbf{Kaixin Li}$^4$  \quad \textbf{Zichen Wen}$^2$ \quad 
    \textbf{Zhong Li}$^5$ \quad 
    \textbf{Conghui He}$^6$ \quad 
    \textbf{Xuming Hu}$^7$ \quad 
    \textbf{Linfeng Zhang}$^{1,2}$\thanks{Corresponding Author (\texttt{\href{mailto:zhanglinfeng@sjtu.edu.cn}{zhanglinfeng@sjtu.edu.cn}}).}
     \\
      $^1$Shanghai Jiao Tong University \quad $^2$EPIC Lab, SJTU \quad  
      $^3$Nanyang Technological University \\ $^4$National University of Singapore \quad 
      $^5$Microsoft Research Asia \quad $^6$Shanghai AI Laboratory \\ $^7$Hong Kong University of Science and Technology (Guangzhou)
}

\begin{document}
\maketitle
\begin{abstract}
Fine-tuning large language models (LLMs) on task-specific data is essential for their effective deployment. As dataset sizes grow, efficiently selecting optimal subsets for training becomes crucial to balancing performance and computational costs. Traditional data selection methods often require fine-tuning a scoring model on the target dataset, which is time-consuming and resource-intensive, or rely on heuristics that fail to fully leverage the model's predictive capabilities. To address these challenges, we propose Data Whisperer, an efficient, training-free, attention-based  method that leverages  few-shot in-context learning with the model to be fine-tuned. Comprehensive evaluations were conducted on both raw and synthetic datasets across diverse tasks and models. Notably, Data Whisperer achieves superior performance compared to the full GSM8K dataset on the Llama-3-8B-Instruct model, using just 10\% of the data, and outperforms existing methods with a 3.1-point improvement and a 7.4$\times$ speedup. The code is available at \url{https://github.com/gszfwsb/Data-Whisperer}.

\end{abstract}

\section{Introduction}
Fine-tuning on task-specific data has become a standard approach for adapting large language models (LLMs) to specialized tasks~\citep{instructgpt}. For instance, \textit{continued pre-training}~\citep{continued_pretrain, chang2024large} involves extending the original pre-training phase on datasets closely aligned with the target domain; \textit{instruction tuning}~\citep{ITsurvey, ustun2024aya} focuses on leveraging instruction-response pairs to improve the model's ability to follow instructions; and \textit{task-specific fine-tuning} enhances model performance on a particular task by using domain-specific data~\citep{S2L, staff}. As datasets continue to expand in size, however, a critical challenge arises: \textit{how to efficiently select the optimal training examples}—a process known as \textit{data selection}—to strike a balance between computational cost and model performance~\citep{ccs, el2n_grand, nuggets, staff, IFD}.

\begin{figure}[tb!]
    \centering
    \includegraphics[width=\linewidth]{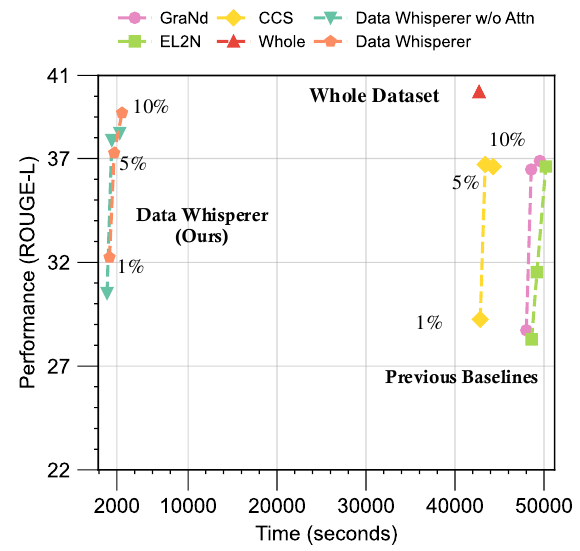}
    \vspace{-20pt}
    \caption{Comparison of the total time and performance across various data selection methods. For each method, we assessed Llama-3-8B-Instruct's performance and time cost when utilizing 1\%, 5\%, and 10\% of the BioInstruct dataset selected by that method. }
    \vspace{-10pt}
    \label{fig:comparison}
\end{figure}

\begin{table*}[tb!]
\centering
\vspace{-10pt}
\caption{Comparison of \textit{Selection-to-Tuning Ratio} (STR) and performance across different selection methods on various datasets using 1\%, 5\%, and 10\% of total data points. Lower STR values indicate higher efficiency. Evaluations were performed using the Llama-3-8B-Instruct model on eight NVIDIA A100 GPUs. \textcolor{PineGreen}{$\uparrow$}/\textcolor{OrangeRed}{$\downarrow$} indicates improvement/degradation compared to random selection. ``Speedup'' represents the acceleration of \mymethod{} over the Nuggets~\citep{nuggets} method. Results for Qwen and Mistral models are provided in Table~\ref{tab:str_appendix}.} 
\vspace{-10pt}
\label{tab:str}
\resizebox{.99\textwidth}{!}{
\begin{tabular}{l|cccc|cccc|cccc}
\toprule
\multirow{3}{*}{\textbf{Method}} & \multicolumn{4}{c|}{\textbf{GSM8K}}& \multicolumn{4}{c|}{\textbf{DialogSum}}& \multicolumn{4}{c}{\textbf{BioInstruct}}\\ 
 & \multicolumn{3}{c}{\textbf{STR}}& \textbf{Performance}& \multicolumn{3}{c}{\textbf{STR}}& \textbf{Performance}& \multicolumn{3}{c}{\textbf{STR}}&\textbf{Performance}\\ 
& 1\% & 5\% & 10\%     &Avg.  Score& 1\% & 5\% & 10\%     &Avg.  Score& 1\% & 5\% & 10\%     &Avg.  Score\\ \midrule
Random  & - & - & -  & 63.77 & - & -& - &  37.66 &- &- &- & 33.89\\
GraNd  & 1.04 & 1.05 & 1.08  & 64.98{\textcolor{PineGreen}{\footnotesize $\uparrow$1.21}} & 1.07 & 1.10 & 1.11  & 36.08{\textcolor{OrangeRed}{\footnotesize $\downarrow$1.58}} & 1.02 & 1.03 & 1.05  & 34.03{\textcolor{PineGreen}{\footnotesize $\uparrow$0.14}} \\
EL2N  &  1.05 & 1.07 & 1.12  & 64.44{\textcolor{PineGreen}{\footnotesize $\uparrow$0.67}} & 1.08 & 1.09 & 1.13  & 32.30{\textcolor{OrangeRed}{\footnotesize $\downarrow$5.36}} & 1.02 & 1.03 & 1.06  & 32.14{\textcolor{OrangeRed}{\footnotesize $\downarrow$1.75}} \\
CCS  & 1.01 & 1.02 & 1.05  & 64.35{\textcolor{PineGreen}{\footnotesize $\uparrow$0.58}} & 1.01 & 1.03 & 1.05  & 30.65{\textcolor{OrangeRed}{\footnotesize $\downarrow$7.01}} & 1.00 & 1.02 & 1.04  & 34.19{\textcolor{PineGreen}{\footnotesize $\uparrow$0.30}} \\
Nuggets  & 1.22 & 1.23 & 1.26  & 66.07{\textcolor{PineGreen}{\footnotesize $\uparrow$2.30}} & 2.42 & 2.46 & 2.53  & 36.25{\textcolor{OrangeRed}{\footnotesize $\downarrow$1.41}} & 0.46 & 0.47 & 0.49  & 32.51{\textcolor{OrangeRed}{\footnotesize $\downarrow$1.38}} \\ 
\mymethod{} & \textbf{0.13} & \textbf{0.14} & \textbf{0.17}  & \textbf{68.14}{\textcolor{PineGreen}{\footnotesize $\uparrow$4.37}} & \textbf{0.12} & \textbf{0.15} & \textbf{0.25}  & \textbf{40.40}{\textcolor{PineGreen}{\footnotesize $\uparrow$2.74}} & \textbf{0.03} & \textbf{0.04} & \textbf{0.06}  & \textbf{36.46}{\textcolor{PineGreen}{\footnotesize $\uparrow$2.57}} \\ \midrule
\rowcolor{mycolor} Speedup & \textbf{9.38$\times$} & \textbf{8.79$\times$} & \textbf{7.41$\times$}  & - & \textbf{20.17$\times$} & \textbf{16.40$\times$} & \textbf{10.12$\times$}  & - & \textbf{15.33$\times$} & \textbf{11.75$\times$} & \textbf{8.17$\times$}  & - \\ \bottomrule
\end{tabular}
}
\vspace{-10pt}
\end{table*}

In this paper, we focus on \textit{data selection for task-specific datasets}, aiming to extract the most informative subset from the original data to achieve nearly lossless performance compared to using the entire dataset for fine-tuning\footnote{We address scenarios where no more advanced LLMs are available beyond the model that is to be fine-tuned, \emph{e.g.}, GPT-4~\citep{gpt4}.}. Although current data selection methods have demonstrated remarkable performance, they often suffer from significant inefficiencies. For instance, as shown in Figure~\ref{fig:comparison}, \textbf{previous state-of-the-art (SOTA) methods can take more time than directly using the entire BioInstruct dataset to fine-tune a Llama-3-8B-Instruct model, even when working with just 1\% of the original dataset.} This inefficiency arises  because these approaches depend on a \textit{fine-tuned} model on the target dataset  for scoring~\citep{IFD,LESS}, typically requiring the same architecture as the model to be fine-tuned. 

To this end, we first critically reevaluate the effectiveness of existing selection methods. To quantitatively and fairly assess the effectiveness of each method, we introduce the \textit{Selection-to-Tuning Ratio} (STR), which is defined as the ratio of time spent on selection to the time required for fine-tuning the model on the entire dataset. Formally, let $t_p(\tau,\rho)$ represent the time associated with a selection method $\tau$ with a budget subset ratio $\rho$, and let $t_{ft}$ denote the corresponding fine-tuning time for the entire dataset. The STR is given by:
\begin{equation}
    \text{STR}(\tau) = \frac{t_p(\tau,\rho)}{t_{ft}}.
    \label{eq:str}
\end{equation}
Intuitively, \textbf{for a data selection method $\tau$ to be considered practically efficient, its STR should ideally be less than 1}. This condition ensures that the time spent on selection aligns with the benefits obtained during fine-tuning. However, as shown in Table~\ref{tab:str}, existing methods often exhibit an STR greater than 1, which constitutes a significant bottleneck in the data selection process, thereby severely limiting the scalability of these methods for large datasets and models.

\begin{figure*}[tb!]
    \centering
    \vspace{-10pt}
    \includegraphics[width=0.9\linewidth]{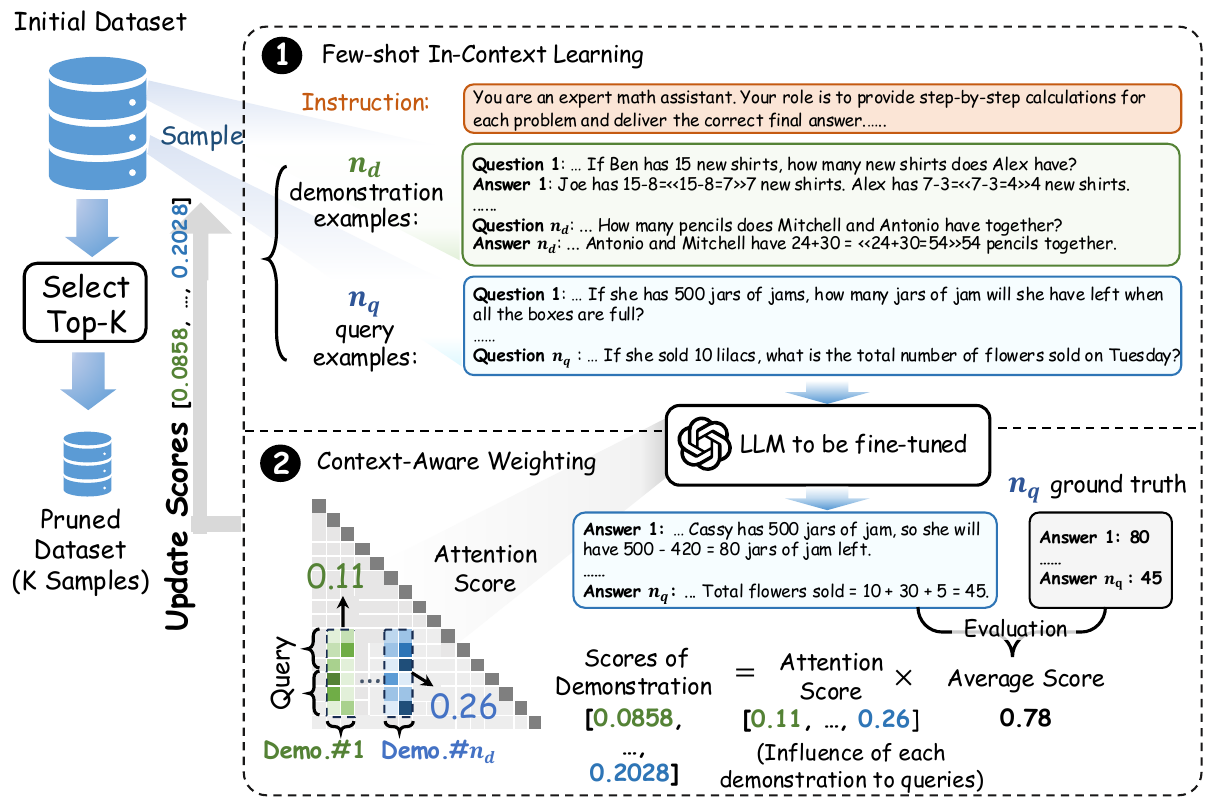}
    \vspace{-10pt}
    \caption{Illustration of the proposed \mymethod{}. $n_d$ and $n_q$ denote the number of demonstrations and queries in ICL. The pipeline consists of two main steps: (I) \textbf{Few-shot In-Context Learning}. A set of demonstration and query examples is randomly sampled from the initial dataset, and an ICL prompt is constructed with a fixed instruction. The LLM to be fine-tuned generates answers for all query examples, and the average evaluation score is computed using the ground truth answers. (II) \textbf{Context-Aware Weighting}. During each iteration of few-shot ICL, we weight the scores of the demonstration examples based on their attention scores, which quantify their influence on the queries. The scores are updated until all samples are scored. The final score for each sample is the average score across its appearances in multiple demonstration sets. Finally, we select the top-k samples from the dataset.}
    \label{fig:method}
    \vspace{-10pt}
\end{figure*}

To address the inefficiencies of existing data selection methods, we introduce \mymethod{}, a training-free, attention-based, and context-aware approach. Traditional methods typically rely on an additional LLM that is fine-tuned on the target dataset to provide scoring for data selection. In contrast, our approach takes advantage of the model's inherent predictive capabilities, inspired by recent theories that equate in-context learning (ICL) with fine-tuning in LLMs~\citep{dai2022can, mosbach2023few}. As shown in Figure~\ref{fig:method}, we score each training sample using few-shot ICL, where the sample itself serves as a demonstration example. The model's performance, measured by its ability to respond to associated queries based on these demonstration examples, yields a raw score for each sample. To improve efficiency, this scoring is performed with a weak-to-strong strategy, \emph{i.e.}, using a less powerful pre-trained model from the same model family.


In typical ICL, \textit{context-awareness} (\emph{e.g.}, the order of examples within the context window) can influence the model’s performance due to its reliance on sequence-based attention mechanisms~\citep{guo-etal-2024-makes,bhope2025optiseq}. To mitigate this inherent order sensitivity, we weight the demonstration scores according to their respective attention scores, as shown in Figure~\ref{fig:method}. This context-aware weighting refines the selection process by ensuring that the final score for each training sample not only reflects its raw performance but also accounts for its contextual significance within the task. The final score for each sample is computed as the average of these weighted scores, which provides a more accurate and nuanced evaluation of each data point’s contribution to the model's learning process.

Our contributions can be summarized as follows:
\begin{enumerate}[leftmargin=10pt, topsep=0pt, itemsep=1pt, partopsep=1pt, parsep=1pt]
\item We critically examine existing data selection approaches and introduce the \textit{Selection-to-Tuning Ratio}, a novel metric that quantifies the efficiency of these methods. We observed that all prior data selection methods are more inefficient than fine-tuning the LLM with the entire dataset.
\item We propose \mymethod{}, an effective, training-free and attention-based method. Unlike previous approaches, our method eliminates the need to fine-tune a separate scoring model on the target dataset, ensuring greater efficiency.
\item \mymethod{} integrates seamlessly with \textit{weak-to-strong} few-shot ICL schemes, enabling effective performance even when a weaker model within the same model family is employed for ICL. This enhances both the scalability and efficiency of our method.
\item Comprehensive experiments are conducted on both real and synthetic datasets across various selection budget ratios, including BioInstruct~\citep{bioinstruct}, DialogSum~\citep{dialogsum}, and GSM8K~\citep{gsm8k}. We observe that \mymethod{} consistently outperforms previous SOTA methods, particularly in smaller data scenarios, while  achieving faster selection times.
\end{enumerate}

\section{Related Work}
\subsection{In-Context Learning}
\textit{In-Context Learning} (ICL) is a powerful task adaptation technique that does not require modifying the weights of a pre-trained model~\citep{brown2020language, olsson2022context, laskin2022context}. Instead, ICL adapts the model to new tasks by conditioning it on a sequence of demonstration pairs, where each demonstration consists of input-output pairs formatted according to a predefined template. This process guides the model in understanding the task. During inference, ICL involves presenting the model with a series of demonstrations followed by a query set, with the model expected to predict the corresponding labels for the query data points based on the context.

Recent studies have explored the theoretical links between ICL and gradient descent, suggesting an implicit relationship between ICL and fine-tuning~\citep{dai2022can, mosbach2023few, deutch2023context, zhou2024context, chen2024bypassing}. One approach, Nuggets~\citep{nuggets}, uses one-shot learning for selection in \textit{instruction-tuning} tasks with log probability scores, but is limited by the computational inefficiency of one-shot learning. In contrast, our work employs attention-based few-shot ICL on the pre-trained model, before any fine-tuning, to directly measure performance scores for \textit{task-specific fine-tuning}. This approach leverages the model's inherent knowledge for data selection, offering a more efficient and scalable solution by linking attention-aware ICL predictions to the fine-tuning process.

\subsection{Data Selection}
Data selection is critical in deep learning, aiming to identify high-quality data that enhances training efficiency while maintaining comparable (or nearly lossless) performance. Traditional methods rely on heuristic metrics, with two primary strategies: (I) \textit{importance}-based selection, which targets challenging or essential samples~\citep{raffel2020exploring, el2n_grand, mirzasoleiman2020coresets, killamsetty2021grad, tan2024data}, and (II) \textit{similarity}- or \textit{diversity}-based selection, which selects samples based on their similarity to others or their representativeness of the feature space~\citep{bukharin2023data, LESS, ccs, wei2021finetuned, yu2024diversify,yang2024clippoweredframeworkrobustgeneralizable,yu2025chainofreasoningunifiedmathematicalreasoning}. For LLMs, these methods face two challenges: (I) the unreliability of heuristic metrics in high-dimensional spaces, and (II) the computational cost of fine-tuning additional scoring models. In contrast, our approach directly integrates data selection with the pre-trained model’s predictions using ICL, eliminating the need for fine-tuning.

\subsection{Data Synthesis}
Data synthesis generates additional high-quality data to enhance model performance and generalization. In LLMs, pretrained models are prompted to produce diverse, context-aware samples that augment scarce or imbalanced datasets~\citep{wang2022self, li2024synthetic, xu2024magpie}, reducing annotation needs and enabling domain adaptation through iterative self-refinement. In vision, data distillation compresses large datasets or models into smaller, representative subsets or synthetic exemplars~\citep{DD,DC,MTT,NCFM,SDC,DRUPI}, cutting redundancy and training costs while maintaining performance for efficient learning.

\section{Method: \mymethod{}}
\subsection{Preliminary: Data Selection in LLM Task-Specific Fine-Tuning}
Given a large and diverse task dataset $\mathcal{D} = \{(x_i,y_i)\}_{i=1}^{|\mathcal{D}|}$, the objective of data selection is to identify a representative subset $\mathcal{D'} \subseteq \mathcal{D}$, with a budget ratio $\rho = |\mathcal{D'}| / |\mathcal{D}| < 1$, for fine-tuning a pre-trained language model $\mathcal{M}_p$. The fine-tuned model, $\mathcal{M}_f$, is then evaluated on the test set $\mathcal{D}_{\textrm{test}}$. The selected subset $\mathcal{D'}$ should ensure that the performance of the fine-tuned model $\mathcal{M}_f(\mathcal{D}')$ remains comparable to that of a model fine-tuned on the entire dataset $\mathcal{M}_f(\mathcal{D})$ when evaluated on $\mathcal{D}_{\textrm{test}}$. The data selection objective is formulated as:
\begin{equation}
    \min_{\mathcal{D'} \subseteq \mathcal{D}, |\mathcal{D'}| = \rho \cdot |\mathcal{D}|} \mathbb{E}_{(x, y) \sim \mathcal{D}_{\text{test}}} \ell(x, y; \mathcal{M}_f(\mathcal{D}')),
\end{equation}
where $\ell(\cdot)$ is the task-specific loss function (\emph{e.g.}, cross-entropy). The key challenge in effective coreset selection is to efficiently identify the coreset $\mathcal{D}'$ for different selection ratios $\rho$.


\subsection{ICL as a Data Selector for LLM Fine-Tuning}
In contrast to conventional methods that rely on handcrafted heuristics of a fine-tuned model $\mathcal{M}_f$, our approach, \mymethod{}, utilizes the intrinsic ICL capabilities of the pre-trained model $\mathcal{M}_p$ for data selection. At each iteration, we randomly sample $n_d$ demonstration examples $\mathcal{D}_d = \{(x_d^{(1)}, y_d^{(1)}), \dots, (x_d^{(n_d)}, y_d^{(n_d)})\}$, and $n_q$ query examples $\mathcal{D}_q = \{x_q^{(1)}, \dots, x_q^{(n_q)}\}$, where $\mathcal{D}_d \cap \mathcal{D}_q = \emptyset$. Combined with a fixed instruction $I$, the total context $C$ is obtained, which is defined as:
\begin{equation}
    C = \{I, (x_d^{(1)}, y_d^{(1)}), \dots, (x_d^{(n_d)}, y_d^{(n_d)})\},
\end{equation}
The whole ICL process is formalized as:
\begin{equation}
    \hat{y}_q^{(1)}, \dots, \hat{y}_q^{(n_q)} = \mathcal{M}_p(C, x_q^{(1)}, \dots, x_q^{(n_q)}).
\end{equation}
Using performance metrics $f$ (\emph{e.g.}, average accuracy, or ROUGE-L~\citep{rouge}, etc.), we compare the predicted outputs $\hat{y}_q^{(1)}, \dots, \hat{y}_q^{(n_q)}$ with the ground truth labels $y_q^{(1)}, \dots, y_q^{(n_q)}$ to compute the average performance score $s$:
\begin{equation}
    s = \frac{1}{n_q}\sum_{j=1}^{n_q} f(\hat{y}_q^{(j)}, y_q^{(j)}).
\end{equation}
This score is assigned to each sample in the demonstration set, and the process is repeated across multiple iterations. The final score for each sample is the average of its scores from all repetitions.

\subsection{Context-Aware Weighting}
To mitigate potential order sensitivity in few-shot ICL, we introduce a \emph{context-aware} weighting mechanism for scoring the demonstration set, as shown in Figure~\ref{fig:method}. This mechanism leverages the self-attention scores from a specific layer $l$ of the pre-trained model $\mathcal{M}_p$, across all attention heads $h$.

We focus on the attention scores corresponding to the first prediction token of a fixed layer $l$. Let $A^{(h)}$ denote the self-attention matrix of layer $l$. For each demonstration example $(x_d^{(i)}, y_d^{(i)})$, we extract the submatrix $A^{(h)}_{(x_d^{(i)}, y_d^{(i)})}$ from $A^{(h)}$, which contains the attention scores between this demonstration and all query examples $x_q^{(1)}, \dots, x_q^{(n_q)}$. The sum of these attention scores quantifies the interaction between the demonstration and the query examples. The weight for each demonstration example, across all attention heads, is computed as:
\begin{equation}
    w_{(x_d^{(i)}, y_d^{(i)})} = \sum_h \mathbf{1}^\top A_{(x_d^{(i)}, y_d^{(i)})}^{(h)} \mathbf{1},
\end{equation}
where $A_{(x_d^{(i)}, y_d^{(i)})}^{(h)}$ is the attention submatrix for head $h$, and the summation is performed across all heads. To account for variations in the length of demonstration examples, we normalize the summed attention scores by the length of each demonstration.

This process is repeated for each demonstration in every iteration until all samples are scored, yielding the scoring set $\mathcal{S}$. If any demonstration example is scored more than once, the scores are averaged. Finally, we select the Top-k samples as follows:
\begin{equation}
    \mathcal{D}' = \text{Top-k}(\mathcal{D};\mathcal{S}),
\end{equation}
where $k = \lfloor \rho \cdot |\mathcal{D}| \rfloor$ denotes the coreset size. The detailed algorithm of \mymethod{} is demonstrated in Algorithm~\ref{alg:method}.

\subsection{Theoretical Analysis}
We now provide a theoretical analysis of \mymethod{} and demonstrate that our approach effectively performs implicit data selection for task-specific fine-tuning. 

\noindent \textbf{Analysis of ICL's Query Prediction}. In ICL, the model adjusts its predictions based on the attention assigned to a set of demonstration examples. Let $x \in \mathbb{R}^{d_{\text{in}}}$ represent the input of a query token $t$, and $q = W_Q x$ denote the attention query vector. The attention result for a specific head is formulated as:
\begin{equation}
\begin{aligned}
     &\mathcal{M}_p(q)=\textrm{Attn}(V,K,q)\\
     &= W_V[X_d;X_{q,<t}]\sigma\left(\frac{(W_K[X_d;X_{q,<t}]^\top q)}{\sqrt{d}}\right),
\end{aligned}
\end{equation}
where $\sigma$ denotes the softmax operator, and $W_Q, W_K, W_V \in \mathbb{R}^{d_{\text{out}} \times d_{\text{in}}}$ are the projection matrices for the attention queries, keys, and values, and $\sqrt{d}$ is the scaling factor. Here, $X_{q,<t}$ represents the input representations of the query tokens before token $t$, $X_d$ represents the input representations of the demonstration tokens, and $[X_d; X_{q,<t}]$ represents the concatenation of these matrices. For qualitative analysis, we approximate the standard attention mechanism by relaxing it to linear attention, removing both the softmax operation and the scaling factor:
\begin{equation}\label{eq:icl_grad}
\begin{aligned}
& \mathcal{M}_p(q)  \approx W_V[X_d;X_{q,<t}] (W_K[X_d;X_{q,<t}])^\top q \\
& = \underbrace{W_V X_{q,<t} (W_K X_{q,<t})^\top q}_{\text{zero shot input}} + \underbrace{W_VX_d (W_K X_d)^\top q}_{\text{ICL examples input}} \\
& = W_{\textrm{zsl}} q + \sum_i(W_Vx_{d}^{(i)})\left((W_K x_{d}^{(i)})^\top q\right)\\
& = W_{\textrm{zsl}} q + \sum_i\left((W_Vx_{d}^{(i)})\otimes(W_K x_{d}^{(i)})\right)q\\
& = (W_{\textrm{zsl}}+\Delta W_{\textrm{icl}})q
\end{aligned}
\end{equation}
In the above derivation, we observe that the attention to the demonstration tokens, $X_d$, can be viewed as an implicit parameter update $\Delta W_{\text{icl}}$, which modifies the zero-shot weights $W_{\text{zsl}}$ to adapt to the task at hand. This update is essentially driven by the in-context learning examples and their relevance to the query tokens.

\begin{table*}[tb!]
\caption{Evaluation results of different data selection methods on the GSM8k, DialogSum, and BioInstruct datasets. The model was fine-tuned on a coreset comprising 1\%, 5\%, and 10\% of the original dataset size. \textcolor{PineGreen}{$\uparrow$} indicates an improvement over random selection, while \textcolor{OrangeRed}{$\downarrow$} indicates a degradation compared to random selection.}
\label{tab:real}
\centering
\vspace{-10pt}
\resizebox{.99\textwidth}{!}{
\begin{tabular}{l|ccc|ccc|ccc}
\toprule
\multirow{2}{*}{\textbf{Method}} & \multicolumn{3}{c|}{\textbf{GSM8k}} & \multicolumn{3}{c|}{\textbf{DialogSum}} & \multicolumn{3}{c}{\textbf{BioInstruct}} \\ 
 & 1\% & 5\% & 10\%  & 1\% & 5\% & 10\%  & 1\% & 5\% & 10\%  \\ \midrule
\textbf{Llama-3-8B-Instruct} (\textit{Zero-shot}) & \multicolumn{3}{c|}{57.49} & \multicolumn{3}{c|}{18.57} & \multicolumn{3}{c}{13.80} \\  
+ Random & 56.95  & 64.71 & 69.66 & 31.66 & 39.88 & 41.45 & 28.38 & 34.61 & 38.70 \\
+ GraNd~\citep{el2n_grand} & 57.23{\textcolor{PineGreen}{\footnotesize $\uparrow$0.28}}& 67.65{\textcolor{PineGreen}{\footnotesize $\uparrow$2.94}} & 70.05{\textcolor{PineGreen}{\footnotesize $\uparrow$0.39}} & 26.02{\textcolor{OrangeRed}{\footnotesize $\downarrow$5.64}} & 40.32{\textcolor{PineGreen}{\footnotesize $\uparrow$0.44}} & 41.90{\textcolor{PineGreen}{\footnotesize $\uparrow$0.45}} & 28.73{\textcolor{PineGreen}{\footnotesize $\uparrow$0.35}} & 36.47{\textcolor{PineGreen}{\footnotesize $\uparrow$1.86}} & 36.88{\textcolor{OrangeRed}{\footnotesize $\downarrow$1.82}} \\
+ EL2N~\citep{el2n_grand} & 60.16{\textcolor{PineGreen}{\footnotesize $\uparrow$3.21}} & 61.50{\textcolor{OrangeRed}{\footnotesize $\downarrow$3.21}} & 71.66{\textcolor{PineGreen}{\footnotesize $\uparrow$2.00}} & 23.05{\textcolor{OrangeRed}{\footnotesize $\downarrow$8.61}} & 35.14{\textcolor{OrangeRed}{\footnotesize $\downarrow$4.74}} & 38.71{\textcolor{OrangeRed}{\footnotesize $\downarrow$2.74}} & 28.29{\textcolor{OrangeRed}{\footnotesize $\downarrow$0.09}} & 31.52{\textcolor{OrangeRed}{\footnotesize $\downarrow$3.09}} & 36.62{\textcolor{OrangeRed}{\footnotesize $\downarrow$2.08}} \\
+ CCS~\citep{ccs} & 60.70{\textcolor{PineGreen}{\footnotesize $\uparrow$3.75}} & 63.64{\textcolor{OrangeRed}{\footnotesize $\downarrow$0.07}} & 68.72{\textcolor{OrangeRed}{\footnotesize $\downarrow$0.94}} & 23.48{\textcolor{OrangeRed}{\footnotesize $\downarrow$8.18}} & 27.34{\textcolor{OrangeRed}{\footnotesize $\downarrow$12.54}} & 41.12{\textcolor{OrangeRed}{\footnotesize $\downarrow$0.33}} & 29.25{\textcolor{PineGreen}{\footnotesize $\uparrow$0.87}} & 36.71{\textcolor{PineGreen}{\footnotesize $\uparrow$2.10}} & 36.61{\textcolor{OrangeRed}{\footnotesize $\downarrow$2.09}} \\
+ Nuggets~\citep{nuggets} & 60.17{\textcolor{PineGreen}{\footnotesize $\uparrow$3.22}} & 68.65{\textcolor{PineGreen}{\footnotesize $\uparrow$3.94}} & 69.39{\textcolor{OrangeRed}{\footnotesize $\downarrow$0.27}} &  31.77{\textcolor{PineGreen}{\footnotesize $\uparrow$0.11}} & 36.85{\textcolor{OrangeRed}{\footnotesize $\downarrow$3.03}} & 40.13{\textcolor{OrangeRed}{\footnotesize $\downarrow$1.32}}  & 22.95{\textcolor{OrangeRed}{\footnotesize $\downarrow$5.43}} & 36.41{\textcolor{PineGreen}{\footnotesize $\uparrow$1.80}} & 38.18{\textcolor{OrangeRed}{\footnotesize $\downarrow$0.52}} \\
\rowcolor{mycolor}+ \textbf{\mymethod{} (ours)} & \textbf{62.57}{\textcolor{PineGreen}{\footnotesize $\uparrow$5.62}} & \textbf{69.65}{\textcolor{PineGreen}{\footnotesize $\uparrow$4.94}} & \textbf{72.46}{\textcolor{PineGreen}{\footnotesize $\uparrow$2.80}} & \textbf{38.05}{\textcolor{PineGreen}{\footnotesize $\uparrow$6.39}} & \textbf{40.96}{\textcolor{PineGreen}{\footnotesize $\uparrow$1.08}} & \textbf{42.18}{\textcolor{PineGreen}{\footnotesize $\uparrow$0.73}} & \textbf{32.29}{\textcolor{PineGreen}{\footnotesize $\uparrow$3.91}} & \textbf{37.27}{\textcolor{PineGreen}{\footnotesize $\uparrow$2.66}} & \textbf{39.20}{\textcolor{PineGreen}{\footnotesize $\uparrow$0.50}} \\ 
\textit{Whole Dataset} & \multicolumn{3}{c|}{71.39} & \multicolumn{3}{c|}{43.33} & \multicolumn{3}{c}{40.21} \\  \midrule 
\textbf{Qwen-2.5-7B-Instruct} (\textit{Zero-shot}) & \multicolumn{3}{c|}{21.23} & \multicolumn{3}{c|}{22.10} & \multicolumn{3}{c}{17.30} \\  
+ Random & 21.51 & 79.28 & 80.08 & 27.31 & 36.68 & 39.99 & 30.65 & 36.62 & 38.30 \\
+ GraNd~\citep{el2n_grand} & 21.98{\textcolor{PineGreen}{\footnotesize $\uparrow$0.47}} & 82.09{\textcolor{PineGreen}{\footnotesize $\uparrow$2.81}} & 83.76{\textcolor{PineGreen}{\footnotesize $\uparrow$3.68}} & 30.61{\textcolor{PineGreen}{\footnotesize $\uparrow$3.30}} & 37.77{\textcolor{PineGreen}{\footnotesize $\uparrow$1.09}} & 40.50{\textcolor{PineGreen}{\footnotesize $\uparrow$0.51}} & 23.70{\textcolor{OrangeRed}{\footnotesize $\downarrow$6.95}} & 35.89{\textcolor{OrangeRed}{\footnotesize $\downarrow$0.73}} & 38.27{\textcolor{OrangeRed}{\footnotesize $\downarrow$0.03}} \\
+ EL2N~\citep{el2n_grand} & 22.13{\textcolor{PineGreen}{\footnotesize $\uparrow$0.62}} & 77.41{\textcolor{OrangeRed}{\footnotesize $\downarrow$1.87}} & 82.62{\textcolor{PineGreen}{\footnotesize $\uparrow$2.54}} & 32.57{\textcolor{PineGreen}{\footnotesize $\uparrow$5.26}} & 38.90{\textcolor{PineGreen}{\footnotesize $\uparrow$2.22}} & 39.45{\textcolor{OrangeRed}{\footnotesize $\downarrow$0.54}} & 30.36{\textcolor{OrangeRed}{\footnotesize $\downarrow$0.29}} & 36.30{\textcolor{OrangeRed}{\footnotesize $\downarrow$0.32}} & 37.26{\textcolor{OrangeRed}{\footnotesize $\downarrow$1.04}} \\
+ CCS~\citep{ccs} & 21.33{\textcolor{OrangeRed}{\footnotesize $\downarrow$0.18}} & 80.75{\textcolor{PineGreen}{\footnotesize $\uparrow$1.47}} & 81.95{\textcolor{PineGreen}{\footnotesize $\uparrow$1.87}} & 31.69{\textcolor{PineGreen}{\footnotesize $\uparrow$4.38}} & 36.15{\textcolor{OrangeRed}{\footnotesize $\downarrow$0.53}} & 38.05{\textcolor{OrangeRed}{\footnotesize $\downarrow$1.94}} & 23.73{\textcolor{OrangeRed}{\footnotesize $\downarrow$6.92}} & 34.53{\textcolor{OrangeRed}{\footnotesize $\downarrow$2.09}} & 37.34{\textcolor{OrangeRed}{\footnotesize $\downarrow$0.96}} \\
+ Nuggets~\citep{nuggets} & 22.70{\textcolor{PineGreen}{\footnotesize $\uparrow$1.19}} & 82.09{\textcolor{PineGreen}{\footnotesize $\uparrow$2.81}} & 83.42{\textcolor{PineGreen}{\footnotesize $\uparrow$3.34}} & 32.11{\textcolor{PineGreen}{\footnotesize $\uparrow$4.80}} & 37.54{\textcolor{PineGreen}{\footnotesize $\uparrow$0.86}} & 40.23{\textcolor{PineGreen}{\footnotesize $\uparrow$0.24}} & 30.25{\textcolor{OrangeRed}{\footnotesize $\downarrow$0.40}} & 36.33{\textcolor{OrangeRed}{\footnotesize $\downarrow$0.29}} & 38.28{\textcolor{OrangeRed}{\footnotesize $\downarrow$0.02}} \\
+ STAFF~\citep{staff} & 22.52{\textcolor{PineGreen}{\footnotesize $\uparrow$1.01}} & 82.22{\textcolor{PineGreen}{\footnotesize $\uparrow$2.94}} & 83.49{\textcolor{PineGreen}{\footnotesize $\uparrow$3.41}} & 30.16{\textcolor{PineGreen}{\footnotesize $\uparrow$2.85}} & 40.84{\textcolor{PineGreen}{\footnotesize $\uparrow$4.16}} & 40.54{\textcolor{PineGreen}{\footnotesize $\uparrow$0.55}} & 23.64{\textcolor{OrangeRed}{\footnotesize $\downarrow$7.01}} & 37.08{\textcolor{PineGreen}{\footnotesize $\uparrow$0.46}} & 38.20{\textcolor{OrangeRed}{\footnotesize $\downarrow$0.10}} \\
\rowcolor{mycolor}+ \textbf{\mymethod{} (ours)} & \textbf{24.45}{\textcolor{PineGreen}{\footnotesize $\uparrow$2.94}} & \textbf{83.16}{\textcolor{PineGreen}{\footnotesize $\uparrow$3.88}} & \textbf{85.03}{\textcolor{PineGreen}{\footnotesize $\uparrow$4.95}} & \textbf{32.95}{\textcolor{PineGreen}{\footnotesize $\uparrow$5.64}} & \textbf{40.95}{\textcolor{PineGreen}{\footnotesize $\uparrow$4.27}} & \textbf{43.00}{\textcolor{PineGreen}{\footnotesize $\uparrow$3.01}} & \textbf{34.93}{\textcolor{PineGreen}{\footnotesize $\uparrow$4.28}} & \textbf{37.57}{\textcolor{PineGreen}{\footnotesize $\uparrow$0.95}} & \textbf{38.85}{\textcolor{PineGreen}{\footnotesize $\uparrow$0.55}} \\ 
\textit{Whole Dataset} & \multicolumn{3}{c|}{85.43} & \multicolumn{3}{c|}{43.79} & \multicolumn{3}{c}{40.71} \\  \midrule 
\textbf{Mistral-Nemo-Instruct} (\textit{Zero-shot}) & \multicolumn{3}{c|}{29.41} & \multicolumn{3}{c|}{19.39} & \multicolumn{3}{c}{13.12} \\ 
+ Random & 32.35 & 54.28 & 64.44 & 19.48 & 37.37 & 40.98 & 13.25 & 24.98 & 36.76 \\
+ GraNd~\citep{el2n_grand} & 31.55{\textcolor{OrangeRed}{\footnotesize $\downarrow$0.80}} & 56.02{\textcolor{PineGreen}{\footnotesize $\uparrow$1.74}} & 67.38{\textcolor{PineGreen}{\footnotesize $\uparrow$2.94}} & 19.86{\textcolor{PineGreen}{\footnotesize $\uparrow$0.38}} & 36.52{\textcolor{OrangeRed}{\footnotesize $\downarrow$0.85}} & 41.79{\textcolor{PineGreen}{\footnotesize $\uparrow$0.81}} & 17.24{\textcolor{PineGreen}{\footnotesize $\uparrow$3.99}} & 25.76{\textcolor{PineGreen}{\footnotesize $\uparrow$0.78}} & 30.05{\textcolor{OrangeRed}{\footnotesize $\downarrow$6.71}} \\
+ EL2N~\citep{el2n_grand} & 30.43{\textcolor{OrangeRed}{\footnotesize $\downarrow$1.92}} & 57.65{\textcolor{PineGreen}{\footnotesize $\uparrow$3.37}} & 67.14{\textcolor{PineGreen}{\footnotesize $\uparrow$2.70}} & 19.97{\textcolor{PineGreen}{\footnotesize $\uparrow$0.49}} & 35.88{\textcolor{OrangeRed}{\footnotesize $\downarrow$1.49}} & 36.40{\textcolor{OrangeRed}{\footnotesize $\downarrow$4.58}} & 14.93{\textcolor{PineGreen}{\footnotesize $\uparrow$1.68}} & 26.20{\textcolor{PineGreen}{\footnotesize $\uparrow$1.22}} & 30.15{\textcolor{OrangeRed}{\footnotesize $\downarrow$6.61}} \\
+ CCS~\citep{ccs} & 32.35{\textcolor{OrangeRed}{\footnotesize $\downarrow$0.00}} & 53.61{\textcolor{OrangeRed}{\footnotesize $\downarrow$0.67}} & 66.18{\textcolor{PineGreen}{\footnotesize $\uparrow$1.74}} & 18.55{\textcolor{OrangeRed}{\footnotesize $\downarrow$0.93}} & 34.57{\textcolor{OrangeRed}{\footnotesize $\downarrow$2.80}} & 42.14{\textcolor{PineGreen}{\footnotesize $\uparrow$1.16}} & 14.42{\textcolor{PineGreen}{\footnotesize $\uparrow$1.17}} & 28.12{\textcolor{PineGreen}{\footnotesize $\uparrow$3.14}} & 36.55{\textcolor{OrangeRed}{\footnotesize $\downarrow$0.21}} \\
+ Nuggets~\citep{nuggets} & 32.09{\textcolor{OrangeRed}{\footnotesize $\downarrow$0.26}} & 58.29{\textcolor{PineGreen}{\footnotesize $\uparrow$4.01}} & 68.79{\textcolor{PineGreen}{\footnotesize $\uparrow$4.35}} & 19.52{\textcolor{PineGreen}{\footnotesize $\uparrow$0.04}} & 37.54{\textcolor{PineGreen}{\footnotesize $\uparrow$0.17}} & 40.52{\textcolor{OrangeRed}{\footnotesize $\downarrow$0.46}} & 20.40{\textcolor{PineGreen}{\footnotesize $\uparrow$7.15}} & 27.40{\textcolor{PineGreen}{\footnotesize $\uparrow$2.42}} & 36.56{\textcolor{OrangeRed}{\footnotesize $\downarrow$0.20}} \\
+ STAFF~\citep{staff} & 31.08{\textcolor{OrangeRed}{\footnotesize $\downarrow$1.27}} & 63.18{\textcolor{PineGreen}{\footnotesize $\uparrow$8.90}} & 67.91{\textcolor{PineGreen}{\footnotesize $\uparrow$3.47}} & 19.49{\textcolor{PineGreen}{\footnotesize $\uparrow$0.01}} & 36.55{\textcolor{OrangeRed}{\footnotesize $\downarrow$0.82}} & 42.08{\textcolor{PineGreen}{\footnotesize $\uparrow$1.10}} & 13.26{\textcolor{PineGreen}{\footnotesize $\uparrow$0.01}} & 25.40{\textcolor{PineGreen}{\footnotesize $\uparrow$0.42}} & 36.44{\textcolor{OrangeRed}{\footnotesize $\downarrow$0.32}} \\
\rowcolor{mycolor}+ \textbf{\mymethod{} (ours)} & \textbf{32.63}{\textcolor{PineGreen}{\footnotesize $\uparrow$0.28}} & \textbf{65.91}{\textcolor{PineGreen}{\footnotesize $\uparrow$11.63}} & \textbf{74.32}{\textcolor{PineGreen}{\footnotesize $\uparrow$9.88}} & \textbf{21.52}{\textcolor{PineGreen}{\footnotesize $\uparrow$2.04}} & \textbf{41.48}{\textcolor{PineGreen}{\footnotesize $\uparrow$4.11}} & \textbf{43.36}{\textcolor{PineGreen}{\footnotesize $\uparrow$2.38}} & \textbf{21.54}{\textcolor{PineGreen}{\footnotesize $\uparrow$8.29}} & \textbf{29.98}{\textcolor{PineGreen}{\footnotesize $\uparrow$5.00}} & \textbf{38.37}{\textcolor{PineGreen}{\footnotesize $\uparrow$1.61}} \\ 
\textit{Whole Dataset} & \multicolumn{3}{c|}{75.00} & \multicolumn{3}{c|}{43.69} & \multicolumn{3}{c}{40.24} \\  \bottomrule 
\end{tabular}
}
\end{table*}

\begin{table*}[tb!]
\centering
\caption{Evaluation results of different data selection methods on the synthetic dataset, generated from DialogSum dataset. The model was fine-tuned on a coreset comprising  5\%,  10\%, and 25\% of the synthetic dataset size.}
\label{tab:synthetic}
\vspace{-10pt}
\resizebox{.99\textwidth}{!}{
\begin{tabular}{l|cccc|cccc|cccc}
\toprule
\multirow{2}{*}{\textbf{Method}} & \multicolumn{4}{c|}{\textbf{Llama-3-8B-Instruct}} & \multicolumn{4}{c|}{\textbf{Qwen-2.5-7B-Instruct}} & \multicolumn{4}{c}{\textbf{Mistral-Nemo-Instruct-2407}} \\
 & 5\% & 10\% & 25\% & 100\% & 5\% & 10\% & 25\% & 100\% & 5\% & 10\% & 25\% & 100\% \\ \midrule
Random & 30.21 & 31.26 & 33.05 & \multirow{6}{*}{35.31} & 26.15 & 27.46 & 31.31 & \multirow{6}{*}{34.55} & 19.24 & 32.44 & 33.09 & \multirow{6}{*}{34.36}\\
GraNd & 27.04{\textcolor{OrangeRed}{\footnotesize $\downarrow$3.17}} & 31.38{\textcolor{PineGreen}{\footnotesize $\uparrow$0.12}} & 33.34{\textcolor{PineGreen}{\footnotesize $\uparrow$0.29}} & & 28.00{\textcolor{PineGreen}{\footnotesize $\uparrow$1.85}} & 29.79{\textcolor{PineGreen}{\footnotesize $\uparrow$2.33}} & 30.59{\textcolor{OrangeRed}{\footnotesize $\downarrow$0.72}} & & 20.03{\textcolor{PineGreen}{\footnotesize $\uparrow$0.79}} & 32.28{\textcolor{OrangeRed}{\footnotesize $\downarrow$0.16}} & 33.05{\textcolor{OrangeRed}{\footnotesize $\downarrow$0.04}} \\
EL2N & 28.67{\textcolor{OrangeRed}{\footnotesize $\downarrow$1.54}} & 31.26{\textcolor{OrangeRed}{\footnotesize $\downarrow$0.00}} & 32.78{\textcolor{OrangeRed}{\footnotesize $\downarrow$0.27}} & & 27.60{\textcolor{PineGreen}{\footnotesize $\uparrow$1.45}} & 27.33{\textcolor{OrangeRed}{\footnotesize $\downarrow$0.13}} & 34.91{\textcolor{PineGreen}{\footnotesize $\uparrow$3.60}} & & 20.96{\textcolor{PineGreen}{\footnotesize $\uparrow$1.72}} & 30.92{\textcolor{OrangeRed}{\footnotesize $\downarrow$1.52}}  & 32.42{\textcolor{OrangeRed}{\footnotesize $\downarrow$0.67}} \\
CCS & 30.19{\textcolor{OrangeRed}{\footnotesize $\downarrow$0.02}} & 32.75{\textcolor{PineGreen}{\footnotesize $\uparrow$1.49}} & 33.77{\textcolor{PineGreen}{\footnotesize $\uparrow$0.72}} & & 26.41{\textcolor{PineGreen}{\footnotesize $\uparrow$0.26}} & 33.59{\textcolor{PineGreen}{\footnotesize $\uparrow$6.13}} & 33.83{\textcolor{PineGreen}{\footnotesize $\uparrow$2.52}} & & 19.77{\textcolor{PineGreen}{\footnotesize $\uparrow$0.53}} & 32.85{\textcolor{PineGreen}{\footnotesize $\uparrow$0.41}} & 34.07{\textcolor{PineGreen}{\footnotesize $\uparrow$0.98}} \\
Nuggets & 30.39{\textcolor{PineGreen}{\footnotesize $\uparrow$0.18}} & 30.83{\textcolor{OrangeRed}{\footnotesize $\downarrow$0.43}} & 33.84{\textcolor{PineGreen}{\footnotesize $\uparrow$0.79}}  & & 28.40{\textcolor{PineGreen}{\footnotesize $\uparrow$2.25}} & 28.02{\textcolor{PineGreen}{\footnotesize $\uparrow$0.56}} & 32.05{\textcolor{PineGreen}{\footnotesize $\uparrow$0.74}} & & 19.98{\textcolor{PineGreen}{\footnotesize $\uparrow$0.74}} & 31.98{\textcolor{OrangeRed}{\footnotesize $\downarrow$0.46}} & 32.46{\textcolor{OrangeRed}{\footnotesize $\downarrow$0.63}}  \\
STAFF & - & - & - & & 27.33{\textcolor{PineGreen}{\footnotesize $\uparrow$1.18}} & 29.64{\textcolor{PineGreen}{\footnotesize $\uparrow$2.18}} & 31.79{\textcolor{PineGreen}{\footnotesize $\uparrow$0.48}} & & 20.88{\textcolor{PineGreen}{\footnotesize $\uparrow$1.64}} & 31.11{\textcolor{OrangeRed}{\footnotesize $\downarrow$1.33}} & 31.33{\textcolor{OrangeRed}{\footnotesize $\downarrow$1.76}}  \\
\rowcolor{mycolor}\textbf{\mymethod{}} & \textbf{32.15}{\textcolor{PineGreen}{\footnotesize $\uparrow$1.94}} & \textbf{32.81} {\textcolor{PineGreen}{\footnotesize $\uparrow$1.55}}& \textbf{34.07}{\textcolor{PineGreen}{\footnotesize $\uparrow$1.02}} & & \textbf{31.27}{\textcolor{PineGreen}{\footnotesize $\uparrow$5.12}} & 
\textbf{34.04}{\textcolor{PineGreen}{\footnotesize $\uparrow$6.58}}&
\textbf{35.20}{\textcolor{PineGreen}{\footnotesize $\uparrow$3.89}} & & \textbf{22.35}{\textcolor{PineGreen}{\footnotesize $\uparrow$3.11}} & \textbf{35.08}{\textcolor{PineGreen}{\footnotesize $\uparrow$2.64}} & \textbf{35.49}{\textcolor{PineGreen}{\footnotesize $\uparrow$2.40}} &\\ \bottomrule
\end{tabular}
}
\end{table*}

\noindent \textbf{Analysis of Fine-Tuning for Query Prediction}. In fine-tuning, model explicitly updates its key and value projections, $W_K$ and $W_V$, through backpropagation to improve task performance. After fine-tuning, the model’s attention can be expressed as:
\begin{equation}\label{eq:ft_grad}
    \begin{aligned}
         \mathcal{M}_f(q)&=(W_V+\Delta W_V) X_{q,<t}  \\ & \cdot X_{q,<t}^\top (W_K+\Delta W_K)^\top q \\
         & = \underbrace{W_V X_{q,<t} (W_K X_{q,<t})^\top q}_{\text{zero shot input}} \\
         &+ \Delta W_V X_{q,<t}(W_K X_{q,<t})^\top q \\
         & + \Delta W_V X_{q,<t} (\Delta W_K X_{q,<t})^\top q \\ 
         &+ \Delta W_V X_{q,<t}(\Delta W_K X_{q,<t})^\top q\\
        & = (W_{\textrm{zsl}}+\Delta W_{\textrm{ft}}) q.
    \end{aligned}
\end{equation}
Here, $\Delta W_K$ and $\Delta W_V$ represent the parameter updates to the key and value projections $W_K$ and $W_V$, respectively, which are learned via backpropagation from task-specific training objectives. The update $\Delta W_{\text{ft}}$ corresponds to the changes in $W_{\text{zsl}}$ introduced by the fine-tuning process.

\noindent \textbf{Connecting ICL Data Selection and Fine-Tuning}. Based on the previous analysis, both ICL and FT follow a similar structure for task adaptation. Both methods modify the attention parameters $W_K$ and $W_V$, but the difference lies in how the model updates them. In ICL, as shown in Eq.~(\ref{eq:icl_grad}), the demonstration examples $x_d^{(i)}$ impact the query prediction by adjusting the attention weights. These weights, $(W_V x_d^{(i)}) \otimes (W_K x_d^{(i)})$, determine relevance of each example.  In contrast, as shown in Eq.~(\ref{eq:ft_grad}), FT explicitly refines $W_K$ and $W_V$ through gradients from the task-specific loss function. 

Given the structural similarity between ICL and FT, we infer that using ICL for data selection is valid for FT. By selecting and weighting the most relevant examples in ICL, we achieve the same performance gains as fine-tuning, but without the need for explicit updates. Thus, \mymethod{} efficiently performs data selection through ICL, identifying the most relevant data points and adapting the model with minimal computational overhead.

\section{Experiments}

\subsection{Experimental Setup}
\noindent \textbf{Datasets.} We conducted experiments across four datasets, including three real-world and one synthetic dataset, each corresponding to a distinct downstream task. Specifically, we used the following datasets: (i) the BioInstruct dataset~\citep{bioinstruct} for biomedical question answering, (ii) the DialogSum dataset~\citep{dialogsum} for dialogue summarization, (iii) GSM8K~\citep{gsm8k} for mathematical reasoning, and (iv) a synthetic variant of DialogSum. Please see Appendix~\ref{sec:ICL_prompt} for detailed prompt designs.

\noindent \textbf{Models.} For each task, we evaluated the performance of three widely-used large language models (LLMs): Llama-3-8B-Instruct~\citep{llama3}, Qwen-2.5-7B-Instruct~\citep{qwen2,qwen2.5}, and Mistral-Nemo-Instruct-2407~\citep{mistral7b}. To explore the potential of using a weaker model for ICL, \mymethod{} also incorporated two smaller LLMs: Qwen-2.5-3B-Instruct~\citep{qwen2,qwen2.5} and Mistral-7B-Instruct-v0.2~\citep{mistral7b}. Notably, since Llama-3-8B-Instruct does not have a smaller variant within its model family, it was excluded from the weak-to-strong analysis.

\noindent \textbf{Baselines.} We compared \mymethod{} with several state-of-the-art data selection techniques: (i) Random Selection, which randomly samples subsets from the dataset, (ii) GraNd~\citep{el2n_grand}, which selects samples based on large gradient norms, (iii) EL2N~\citep{el2n_grand}, which prioritizes samples with large discrepancies between model predictions and ground truth, (iv) CCS~\citep{ccs}, which balances data coverage and importance, (v) Nuggets~\citep{nuggets}, which employs one-shot learning of a fine-tuned model to identify high-quality examples, and (vi) STAFF~\citep{staff}, which estimates gradient effort scores with a smaller model and validates them on the target LLM to allocate budget across different model regions.

\noindent \textbf{Evaluation.} All experiments were performed on 8 NVIDIA A100 GPUs. For model fine-tuning, we utilized LoRA~\citep{lora} and performed 3 times. Performance on the BioInstruct and DialogSum datasets was assessed using the ROUGE-L metric~\citep{rouge} during both the data selection and fine-tuning stages. For the GSM8K dataset, we computed performance based on the Exact Match (EM) metric by comparing the model's generated answers against the ground truth.

\noindent \textbf{Synthetic Data Generation.} We generated a synthetic variant of the DialogSum dataset using the Llama-3-8B-Instruct model. For every five demonstration samples from the original dataset, we prompted the model to generate one synthetic sample, ensuring alignment with the original dialogue structure. The generated samples were manually reviewed and filtered to remove incoherent, factually incorrect, or misformatted data. See Appendix~\ref{sec:synthetic} for details on synthetic data generation.

\begin{table}[tb!]
    \centering
    \vspace{-5pt}
    \caption{Ablation study on the number of demonstrations $n_d$ and queries $n_q$ in \mymethod{}. We observed that selecting a moderate number of $n_q$ and $n_d$ often yields a better performance for \mymethod{}.}
    \label{tab:ablation_demo_query}
    \vspace{-8pt}
    \resizebox{.49\textwidth}{!}{
    \begin{tabular}{cc|ccc|ccc}
    \toprule 
         \multirow{3}{*}{Dataset}&  \multirow{3}{*}{Ratio} & \multicolumn{3}{c|}{Llama-3-8B-Instruct}&  \multicolumn{3}{c}{Qwen-2.5-7B-Instruct}\\ 
          & &  \makecell{$n_d=5$ \\ $n_q=3$} &  \makecell{$10$ \\ $5$} &  \makecell{$15$ \\ $10$} &  \makecell{$5$ \\ $3$} &  \makecell{$10$ \\ $5$} &  \makecell{$15$ \\ $10$}\\ \midrule
         \multirow{4}{*}{GSM8K}&0\%&   & 57.49 & &  &21.23&\\ 
          &1\%&  \textbf{62.57} &  60.96 &  61.23 &  22.43 &  \textbf{24.45} &  23.75 \\
          &5\%&  68.18 &  \textbf{69.65} &  66.56 &  81.95 &  \textbf{83.01} &  80.95 \\
          &10\%&  70.95 &  \textbf{72.46} &  70.72 &  83.75 &  84.89 &  \textbf{85.03} \\
          \midrule
         \multirow{4}{*}{DialogSum} &0\%& &18.57&& &22.10&\\
          &1\%& 36.11 & \textbf{36.47} & 36.45 & \textbf{ 32.49 } &32.20& 32.14 \\
          &5\%& 38.15 & \textbf{40.96} & 38.43 & 38.02 & \textbf{40.95} & 38.46 \\
          &10\%& 41.27 & \textbf{42.18} & 41.42 & 37.88 & \textbf{43.00} & 40.31 \\
          \midrule
         \multirow{4}{*}{BioInstruct} &0\%& &13.80& & &17.30&\\
          &1\%& \textbf{32.29} & 32.24 & 30.45 & 32.34 & \textbf{34.93} & 32.07 \\
          &5\%& 32.96 & \textbf{37.27} & 34.01 & 36.74 & \textbf{37.14} & 36.64 \\
          &10\%& 38.98 & \textbf{39.20} & 37.67 & 38.20 & \textbf{38.85} & 38.16 \\
        \bottomrule
            \end{tabular}
    }
    \vspace{-5pt}
\end{table}

\subsection{Main Results and Key Observations}

\noindent \textbf{$\bullet$ Smaller Dataset, Comparable Performance with the Whole Dataset.} For real datasets, as shown in Table~\ref{tab:real}, results demonstrate that \mymethod{} performs remarkably well on smaller datasets, achieving comparable performance to full dataset across various selection ratios. For instance, on the GSM8K dataset, fine-tuning with 10\% of data yields even better performance to the model fine-tuned on the entire dataset. For synthetic dataset in Table~\ref{tab:synthetic}, two out of three models outperform the models fine-tuned on the entire dataset.

\noindent \textbf{$\bullet$ Same Size, Better Performance with SOTA Baseline Methods.} Compared to SOTA methods, \mymethod{} demonstrates consistent superiority across varying dataset sizes. On real datasets, as illustrated in Table~\ref{tab:real}, \mymethod{} achieves higher accuracy. For instance, on 10\% data of DialogSum with Qwen-2.5-7B-Instruct, \mymethod{} attains an accuracy of 43.00, surpassing the previous SOTA method, STAFF, by a significant margin of 2.46. Similarly, on synthetic datasets, as shown in Table~\ref{tab:synthetic}, \mymethod{} consistently delivers the best performance across all evaluated models and data proportions, underscoring its robust generalization capabilities. Notably, with the Qwen-2.5-7B-Instruct model on 5\% of the data, \mymethod{} achieves an accuracy of 31.27, outperforming the prior SOTA method, Nuggets, by a remarkable 2.87 points.

\begin{table}[tb!]
\vspace{-5pt}
\caption{Results of different attention layers for context-aware weighting, and the impact of context-aware weighting (w/ vs. w/o) on Llama-3-8B-Instruct.}
\vspace{-8pt}
\centering
\label{tab:layer_ablation}
\resizebox{.49\textwidth}{!}{
\begin{tabular}{@{}cc|ccccc@{}}
\toprule
\multirow{2}{*}{Dataset} & \multirow{2}{*}{Ratio} &   \multirow{2}{*}{Random}&\multicolumn{4}{c}{\mymethod{}}\\  
\cmidrule{4-7}
 &  &   &w/o Attn&Shallow& Intermediate& Deep\\ \midrule
\multirow{4}{*}{GSM8K} & 0\% & \multicolumn{5}{c}{57.49}\\
 & 1\% &   56.95 &60.56&60.73& 60.96 & \textbf{62.30} \\
 & 5\% &   64.71 &68.45 & 68.98 & \textbf{69.65} & 68.71 \\
 & 10\% &   69.66 &71.72 & 71.25 & \textbf{72.46} & 71.79 \\
 \midrule
\multirow{4}{*}{DialogSum} & 0\% &  \multicolumn{5}{c}{18.57}\\
 & 1\% &   31.66 &34.35  & 37.97  & 36.47  & \textbf{38.05}  \\
 & 5\% &   39.88 &40.29 &39.96& \textbf{40.96}  & 40.08\\
 & 10\% &   41.45 &41.91  &41.94 & \textbf{42.18}  & 41.80 \\
 \midrule
\multirow{4}{*}{BioInstruct} & 0\% &\multicolumn{5}{c}{13.80}\\ 
 & 1\% &   28.38 &30.51 &29.27 & \textbf{32.24} & 29.44 \\
 & 5\% &  34.61 &36.38 &36.36 & \textbf{37.27}  & 36.32 \\
 & 10\% &   38.70 &38.83 &38.86 & \textbf{39.20} & 38.79\\
 
 \bottomrule
\end{tabular}
}
\vspace{-5pt}
\end{table}

\subsection{Ablation Study}

\noindent \textbf{$\bullet$ Sensitivity of demonstration and query numbers $n_d$ and $n_q$}. An interesting question is how the number of demonstration examples ($n_d$) and query examples ($n_q$) affects the performance of \mymethod{}. To explore this, we varied the values of $n_d$ and $n_q$ and observed their impact across all three datasets. As shown in Table~\ref{tab:ablation_demo_query}, results indicate that while increasing $n_d$ or $n_q$ improves the model's performance at first, the effect tends to plateau after a certain threshold. This suggests that there exists an optimal balance between the number of demonstration and query examples, beyond which the improvements are marginal, demonstrating the robustness of \mymethod{}. We used $n_d=10$ and $n_q=5$ by default.

\begin{figure}[tb!]
    \centering
    \vspace{-10pt}
    \includegraphics[width=0.99\linewidth]{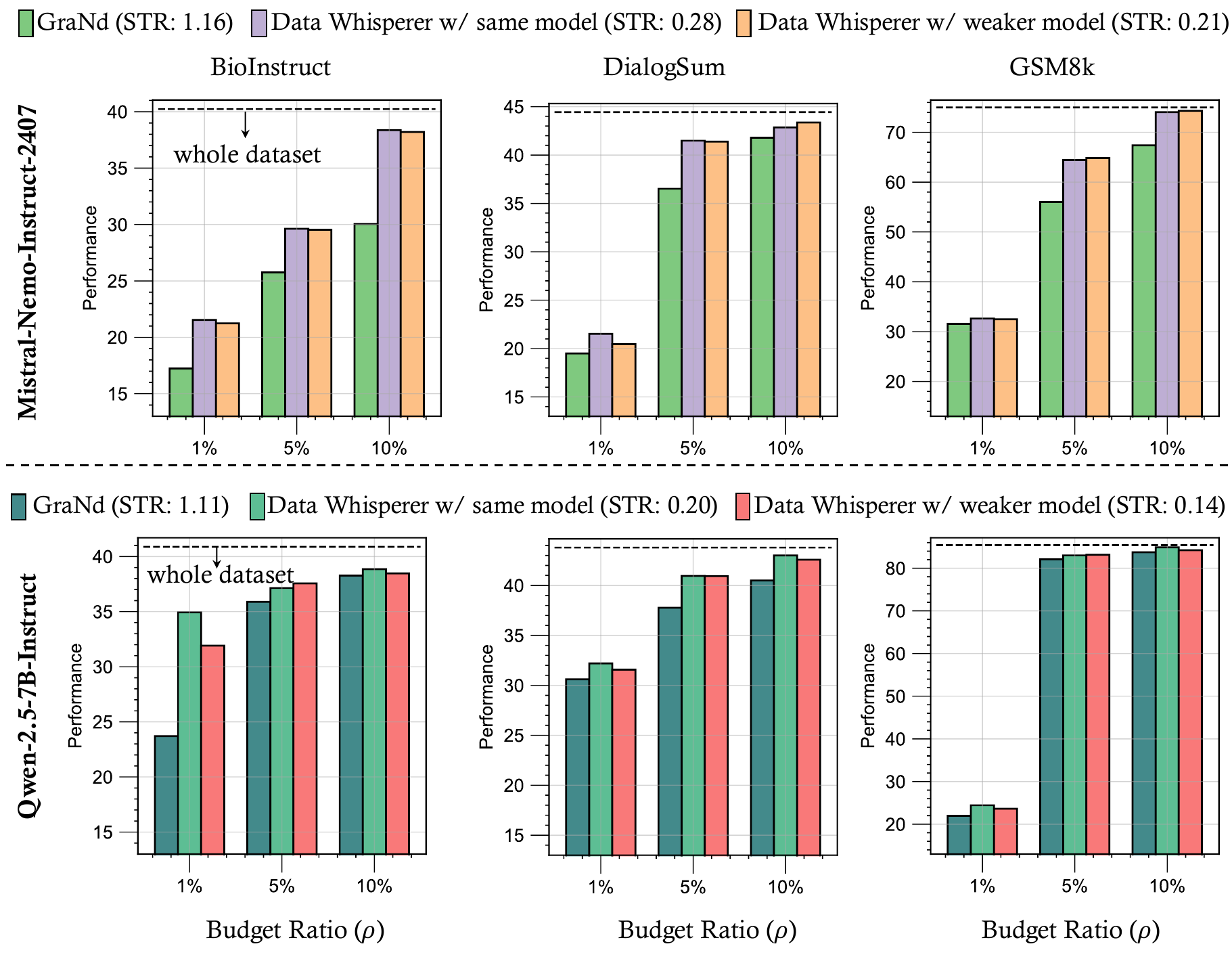}
    \vspace{-25pt}
    \caption{Ablation on the weak-to-strong scalability. For Qwen models, we used Qwen-2.5-3B-Instruct to perform \mymethod{} for Qwen-2.5-7B-Instruct. For Mistral models, we used Mistral-7B-Instruct-v0.2 to perform \mymethod{} for Mistral-Nemo-Instruct-2407. Weaker models were not fine-tuned on the task dataset. We found that using a weaker model does not significantly affect performance and provides a more efficient solution (measured by STR). \textbf{Best viewed in color.}}
    \vspace{-15pt}
    \label{fig:w2s}
\end{figure}

\noindent \textbf{$\bullet$ Ablation of Attention Layers}. The effectiveness of the context-aware weighting mechanism depends on the selected attention layer for scoring. We conducted an ablation study on the Llama-3-8B-Instruct model, evaluating shallow (Layer 5), intermediate (Layer 13), and deep (Layer 30) layers. As shown in Table~\ref{tab:layer_ablation}, the intermediate layer (\emph{e.g.}, Layer 13) generally yielded more stable results, though all layers outperformed random selection. This suggests that \mymethod{} benefits from multiple attention layers, with intermediate layers providing more relevant contextual information for task-specific fine-tuning. These findings align with previous research showing that \textbf{intermediate layers are crucial for semantic interpretation, processing moderately complex concepts, and providing contextual information for tasks requiring understanding}~\citep{wendler2024llamas,jin2025exploring,li2024safety}.

\noindent \textbf{$\bullet$ Effect of weak-to-strong scoring}. We also investigated the impact of weak-to-strong scoring, where a weaker model is used to select data for fine-tuning a stronger model. As shown in Figure~\ref{fig:w2s}, results indicate that \textbf{using a weaker model does not significantly impact the overall performance of \mymethod{}, while providing a more efficient solution with a lower STR}. It demonstrates that \mymethod{} is scalable across different model sizes and highlights its potential for efficient fine-tuning, even with limited computational resources.

\begin{figure}[tb!]
    \centering
    \vspace{-10pt}
    \includegraphics[width=0.99\linewidth]{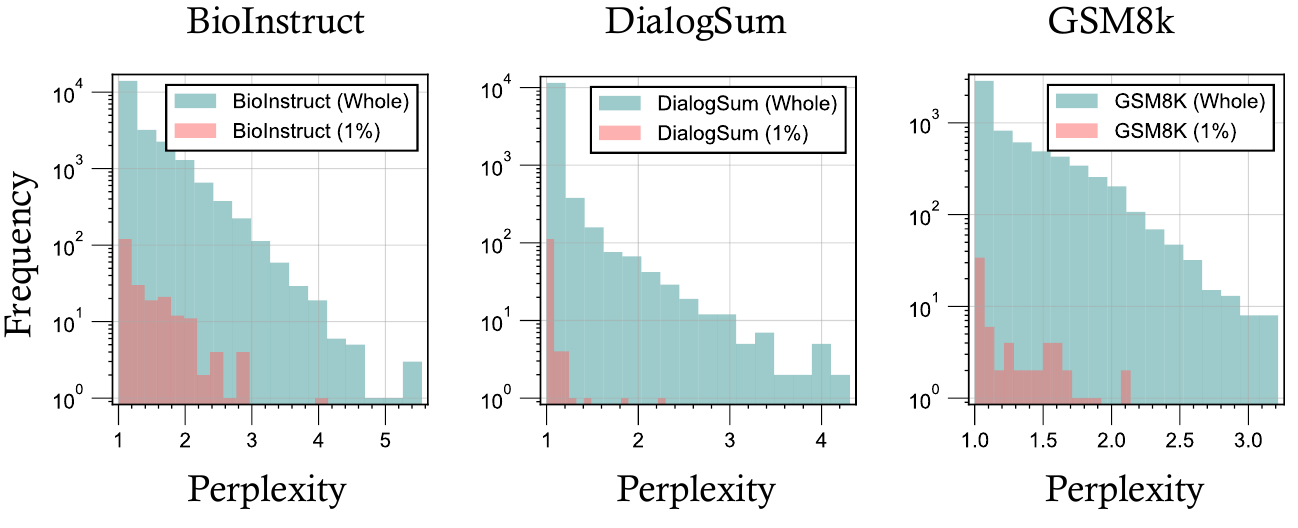}
    \vspace{-25pt}
    \caption{Perplexity distribution of all samples and the selected 1\% samples by \mymethod{}. Perplexity scores are calculated using the GPT-4o-mini model API. The results show that \mymethod{} prefers selecting easier samples, which aligns with the theory in~\citep{sorscher2022beyond} that suggests selecting the easiest samples is optimal in small data scenarios (\emph{e.g.}, 1\% data).}
    \vspace{-8pt}
    \label{fig:ppl}
\end{figure}

\section{Discussion: What Samples Do Task-Specific Fine-Tuning Prefer?}
To study what kinds of samples task-specific fine-tuning favors, we analyzed the perplexity of both the full dataset and the selected 1\% subset across three different task datasets, including BioInstruct, GSM8K, and DialogSum. Specifically, we used GPT-4o mini~\citep{gpt4} to score the perplexity of both the full data and the 1\% selected subset. As shown in Figure~\ref{fig:ppl}, our findings show that the selected 1\% of data consistently exhibits lower perplexity (\emph{i.e.}, easier data) across all three datasets. This aligns with previous findings in \citep{sorscher2022beyond}, which suggest that, \textbf{in small data scenarios, the model tends to prefer easier data to boost performance}. Lower perplexity data tends to be more predictable and less ambiguous, allowing the model to learn task-specific patterns more efficiently when fine-tuned on such samples. This observation further supports the notion that task-specific fine-tuning benefits from focusing on simpler, high-confidence examples, especially in data-limited settings.

\section{Conclusion}
In this paper, we reevaluate existing data selection methods through the lens of the Selection-to-Tuning Ratio, revealing that many traditional approaches fall short in practical scenarios. To address this gap, \mymethod{} introduces an efficient, training-free, attention-based approach to data selection. By leveraging the theoretical connection between ICL and fine-tuning, \mymethod{} eliminates the need for fine-tuning an additional LLM on the target dataset for scoring. Specifically, we use few-shot ICL for data selection, with randomly sampled demonstration and query examples drawn from the initial dataset. Moreover, to mitigate the inherent order sensitivity in ICL, our method incorporates a context-aware weighting strategy based on attention scores. In conclusion, \mymethod{} provides a scalable, efficient solution for data selection in task-specific LLM fine-tuning, offering significant potential for enhancing both efficiency and performance.


\section{Limitation}
While \mymethod{} demonstrates strong performance across a range of tasks and datasets, there are opportunities for further exploration. The attention-based scoring mechanism relies on access to the internal attention layers of the LLM, which may limit its direct applicability in certain settings where such access is unavailable, such as in some commercial API-driven models (\emph{e.g.}, GPT-4~\cite{gpt4} and DeepSeek-R1~\cite{deepseekr1}). Additionally, while our evaluations provide a solid foundation, further research is needed to explore the performance of \mymethod{} with larger LLMs and datasets. Expanding experiments to include larger and more complex models would offer valuable insights into the scalability of the method and its potential in extreme-scale scenarios. Addressing these areas in future work will help maximize the versatility and impact of \mymethod{} across a broader range of applications.

\bibliography{custom}

\begin{thebibliography}{56}
\providecommand{\natexlab}[1]{#1}

\bibitem[{Bhope et~al.(2025)Bhope, Venkateswaran, Jayaram, Isahagian, Muthusamy, and Venkatasubramanian}]{bhope2025optiseq}
Rahul~Atul Bhope, Praveen Venkateswaran, KR~Jayaram, Vatche Isahagian, Vinod Muthusamy, and Nalini Venkatasubramanian. 2025.
\newblock Optiseq: Optimizing example ordering for in-context learning.
\newblock \emph{arXiv preprint arXiv:2501.15030}.

\bibitem[{Brown et~al.(2020)Brown, Mann, Ryder, Subbiah, Kaplan, Dhariwal, Neelakantan, Shyam, Sastry, Askell et~al.}]{brown2020language}
Tom Brown, Benjamin Mann, Nick Ryder, Melanie Subbiah, Jared~D Kaplan, Prafulla Dhariwal, Arvind Neelakantan, Pranav Shyam, Girish Sastry, Amanda Askell, et~al. 2020.
\newblock Language models are few-shot learners.
\newblock \emph{Advances in neural information processing systems}, 33:1877--1901.

\bibitem[{Bukharin and Zhao(2023)}]{bukharin2023data}
Alexander Bukharin and Tuo Zhao. 2023.
\newblock Data diversity matters for robust instruction tuning.
\newblock \emph{arXiv preprint arXiv:2311.14736}.

\bibitem[{Cazenavette et~al.(2022)Cazenavette, Wang, Torralba, Efros, and Zhu}]{MTT}
George Cazenavette, Tongzhou Wang, Antonio Torralba, Alexei~A. Efros, and Jun-Yan Zhu. 2022.
\newblock Dataset distillation by matching training trajectories.
\newblock In \emph{Proceedings of the IEEE/CVF Conference on Computer Vision and Pattern Recognition}.

\bibitem[{Chang et~al.(2024)Chang, Park, Ye, Yang, Seo, Chang, and Seo}]{chang2024large}
Hoyeon Chang, Jinho Park, Seonghyeon Ye, Sohee Yang, Youngkyung Seo, Du-Seong Chang, and Minjoon Seo. 2024.
\newblock How do large language models acquire factual knowledge during pretraining?
\newblock \emph{arXiv preprint arXiv:2406.11813}.

\bibitem[{Chen et~al.(2024)Chen, Li, Liang, Shi, and Song}]{chen2024bypassing}
Bo~Chen, Xiaoyu Li, Yingyu Liang, Zhenmei Shi, and Zhao Song. 2024.
\newblock Bypassing the exponential dependency: Looped transformers efficiently learn in-context by multi-step gradient descent.
\newblock \emph{arXiv preprint arXiv:2410.11268}.

\bibitem[{Chen et~al.(2021)Chen, Liu, Chen, and Zhang}]{dialogsum}
Yulong Chen, Yang Liu, Liang Chen, and Yue Zhang. 2021.
\newblock \href {https://doi.org/10.18653/v1/2021.findings-acl.449} {{D}ialog{S}um: {A} real-life scenario dialogue summarization dataset}.
\newblock In \emph{Findings of the Association for Computational Linguistics: ACL-IJCNLP 2021}, pages 5062--5074, Online. Association for Computational Linguistics.

\bibitem[{Cobbe et~al.(2021)Cobbe, Kosaraju, Bavarian, Chen, Jun, Kaiser, Plappert, Tworek, Hilton, Nakano, Hesse, and Schulman}]{gsm8k}
Karl Cobbe, Vineet Kosaraju, Mohammad Bavarian, Mark Chen, Heewoo Jun, Lukasz Kaiser, Matthias Plappert, Jerry Tworek, Jacob Hilton, Reiichiro Nakano, Christopher Hesse, and John Schulman. 2021.
\newblock Training verifiers to solve math word problems.
\newblock \emph{arXiv preprint arXiv:2110.14168}.

\bibitem[{Dai et~al.(2022)Dai, Sun, Dong, Hao, Ma, Sui, and Wei}]{dai2022can}
Damai Dai, Yutao Sun, Li~Dong, Yaru Hao, Shuming Ma, Zhifang Sui, and Furu Wei. 2022.
\newblock Why can gpt learn in-context? language models implicitly perform gradient descent as meta-optimizers.
\newblock \emph{arXiv preprint arXiv:2212.10559}.

\bibitem[{Deutch et~al.(2023)Deutch, Magar, Natan, and Dar}]{deutch2023context}
Gilad Deutch, Nadav Magar, Tomer~Bar Natan, and Guy Dar. 2023.
\newblock In-context learning and gradient descent revisited.
\newblock \emph{arXiv preprint arXiv:2311.07772}.

\bibitem[{Dubey et~al.(2024)Dubey, Jauhri, Pandey, Kadian, Al-Dahle, Letman, Mathur, Schelten, Yang, Fan et~al.}]{llama3}
Abhimanyu Dubey, Abhinav Jauhri, Abhinav Pandey, Abhishek Kadian, Ahmad Al-Dahle, Aiesha Letman, Akhil Mathur, Alan Schelten, Amy Yang, Angela Fan, et~al. 2024.
\newblock The llama 3 herd of models.
\newblock \emph{arXiv preprint arXiv:2407.21783}.

\bibitem[{Guo et~al.(2025)Guo, Yang, Zhang, Song, Zhang, Xu, Zhu, Ma, Wang, Bi et~al.}]{deepseekr1}
Daya Guo, Dejian Yang, Haowei Zhang, Junxiao Song, Ruoyu Zhang, Runxin Xu, Qihao Zhu, Shirong Ma, Peiyi Wang, Xiao Bi, et~al. 2025.
\newblock Deepseek-r1: Incentivizing reasoning capability in llms via reinforcement learning.
\newblock \emph{arXiv preprint arXiv:2501.12948}.

\bibitem[{Guo et~al.(2024)Guo, Wang, Wang, Ye, and Zhang}]{guo-etal-2024-makes}
Qi~Guo, Leiyu Wang, Yidong Wang, Wei Ye, and Shikun Zhang. 2024.
\newblock \href {https://doi.org/10.18653/v1/2024.findings-acl.884} {What makes a good order of examples in in-context learning}.
\newblock In \emph{Findings of the Association for Computational Linguistics: ACL 2024}, pages 14892--14904, Bangkok, Thailand. Association for Computational Linguistics.

\bibitem[{Gururangan et~al.(2020)Gururangan, Marasovi{\'c}, Swayamdipta, Lo, Beltagy, Downey, and Smith}]{continued_pretrain}
Suchin Gururangan, Ana Marasovi{\'c}, Swabha Swayamdipta, Kyle Lo, Iz~Beltagy, Doug Downey, and Noah~A. Smith. 2020.
\newblock \href {https://doi.org/10.18653/v1/2020.acl-main.740} {Don`t stop pretraining: Adapt language models to domains and tasks}.
\newblock In \emph{Proceedings of the 58th Annual Meeting of the Association for Computational Linguistics}, pages 8342--8360, Online. Association for Computational Linguistics.

\bibitem[{Hu et~al.(2021)Hu, Shen, Wallis, Allen-Zhu, Li, Wang, Wang, and Chen}]{lora}
Edward~J. Hu, Yelong Shen, Phillip Wallis, Zeyuan Allen-Zhu, Yuanzhi Li, Shean Wang, Lu~Wang, and Weizhu Chen. 2021.
\newblock \href {https://arxiv.org/abs/2106.09685} {Lora: Low-rank adaptation of large language models}.
\newblock \emph{Preprint}, arXiv:2106.09685.

\bibitem[{Jiang et~al.(2023)Jiang, Sablayrolles, Mensch, Bamford, Chaplot, Casas, Bressand, Lengyel, Lample, Saulnier et~al.}]{mistral7b}
Albert~Q Jiang, Alexandre Sablayrolles, Arthur Mensch, Chris Bamford, Devendra~Singh Chaplot, Diego de~las Casas, Florian Bressand, Gianna Lengyel, Guillaume Lample, Lucile Saulnier, et~al. 2023.
\newblock Mistral 7b.
\newblock \emph{arXiv preprint arXiv:2310.06825}.

\bibitem[{Jin et~al.(2025)Jin, Yu, Huang, Zeng, Wang, Hua, Zhao, Mei, Meng, Ding et~al.}]{jin2025exploring}
Mingyu Jin, Qinkai Yu, Jingyuan Huang, Qingcheng Zeng, Zhenting Wang, Wenyue Hua, Haiyan Zhao, Kai Mei, Yanda Meng, Kaize Ding, et~al. 2025.
\newblock Exploring concept depth: How large language models acquire knowledge and concept at different layers?
\newblock In \emph{Proceedings of the 31st International Conference on Computational Linguistics}, pages 558--573.

\bibitem[{Killamsetty et~al.(2021)Killamsetty, Durga, Ramakrishnan, De, and Iyer}]{killamsetty2021grad}
Krishnateja Killamsetty, Sivasubramanian Durga, Ganesh Ramakrishnan, Abir De, and Rishabh Iyer. 2021.
\newblock Grad-match: Gradient matching based data subset selection for efficient deep model training.
\newblock In \emph{International Conference on Machine Learning}, pages 5464--5474. PMLR.

\bibitem[{Laskin et~al.(2022)Laskin, Wang, Oh, Parisotto, Spencer, Steigerwald, Strouse, Hansen, Filos, Brooks et~al.}]{laskin2022context}
Michael Laskin, Luyu Wang, Junhyuk Oh, Emilio Parisotto, Stephen Spencer, Richie Steigerwald, DJ~Strouse, Steven Hansen, Angelos Filos, Ethan Brooks, et~al. 2022.
\newblock In-context reinforcement learning with algorithm distillation.
\newblock \emph{arXiv preprint arXiv:2210.14215}.

\bibitem[{Li et~al.(2024{\natexlab{a}})Li, Dong, Tang, Wang, Zhang, Huang, Huang, Huang, Huang, Zhang et~al.}]{li2024synthetic}
Haoran Li, Qingxiu Dong, Zhengyang Tang, Chaojun Wang, Xingxing Zhang, Haoyang Huang, Shaohan Huang, Xiaolong Huang, Zeqiang Huang, Dongdong Zhang, et~al. 2024{\natexlab{a}}.
\newblock Synthetic data (almost) from scratch: Generalized instruction tuning for language models.
\newblock \emph{arXiv preprint arXiv:2402.13064}.

\bibitem[{Li et~al.(2023{\natexlab{a}})Li, Zhang, Li, Chen, Chen, Cheng, Wang, Zhou, and Xiao}]{IFD}
Ming Li, Yong Zhang, Zhitao Li, Jiuhai Chen, Lichang Chen, Ning Cheng, Jianzong Wang, Tianyi Zhou, and Jing Xiao. 2023{\natexlab{a}}.
\newblock From quantity to quality: Boosting llm performance with self-guided data selection for instruction tuning.
\newblock \emph{arXiv preprint arXiv:2308.12032}.

\bibitem[{Li et~al.(2024{\natexlab{b}})Li, Yao, Zhang, and Li}]{li2024safety}
Shen Li, Liuyi Yao, Lan Zhang, and Yaliang Li. 2024{\natexlab{b}}.
\newblock Safety layers in aligned large language models: The key to llm security.
\newblock \emph{arXiv preprint arXiv:2408.17003}.

\bibitem[{Li et~al.(2023{\natexlab{b}})Li, Hui, Xia, Yang, Yang, Zhang, Si, Liu, Liu, Huang et~al.}]{nuggets}
Yunshui Li, Binyuan Hui, Xiaobo Xia, Jiaxi Yang, Min Yang, Lei Zhang, Shuzheng Si, Junhao Liu, Tongliang Liu, Fei Huang, et~al. 2023{\natexlab{b}}.
\newblock One shot learning as instruction data prospector for large language models.
\newblock \emph{arXiv preprint arXiv:2312.10302}.

\bibitem[{Lin(2004)}]{rouge}
Chin-Yew Lin. 2004.
\newblock \href {https://www.aclweb.org/anthology/W04-1013} {{ROUGE}: A package for automatic evaluation of summaries}.
\newblock In \emph{Text Summarization Branches Out}, pages 74--81, Barcelona, Spain. Association for Computational Linguistics.

\bibitem[{Mirzasoleiman et~al.(2020)Mirzasoleiman, Bilmes, and Leskovec}]{mirzasoleiman2020coresets}
Baharan Mirzasoleiman, Jeff Bilmes, and Jure Leskovec. 2020.
\newblock Coresets for data-efficient training of machine learning models.
\newblock In \emph{International Conference on Machine Learning}, pages 6950--6960. PMLR.

\bibitem[{Mosbach et~al.(2023)Mosbach, Pimentel, Ravfogel, Klakow, and Elazar}]{mosbach2023few}
Marius Mosbach, Tiago Pimentel, Shauli Ravfogel, Dietrich Klakow, and Yanai Elazar. 2023.
\newblock Few-shot fine-tuning vs. in-context learning: A fair comparison and evaluation.
\newblock \emph{arXiv preprint arXiv:2305.16938}.

\bibitem[{Olsson et~al.(2022)Olsson, Elhage, Nanda, Joseph, DasSarma, Henighan, Mann, Askell, Bai, Chen et~al.}]{olsson2022context}
Catherine Olsson, Nelson Elhage, Neel Nanda, Nicholas Joseph, Nova DasSarma, Tom Henighan, Ben Mann, Amanda Askell, Yuntao Bai, Anna Chen, et~al. 2022.
\newblock In-context learning and induction heads.
\newblock \emph{arXiv preprint arXiv:2209.11895}.

\bibitem[{OpenAI et~al.(2023)OpenAI, Achiam, Adler, Agarwal, Ahmad, Akkaya, Aleman, Almeida, Altenschmidt, Altman, Anadkat et~al.}]{gpt4}
OpenAI, Josh Achiam, Steven Adler, Sandhini Agarwal, Lama Ahmad, Ilge Akkaya, Florencia~Leoni Aleman, Diogo Almeida, Janko Altenschmidt, Sam Altman, Shyamal Anadkat, et~al. 2023.
\newblock Gpt-4 technical report.
\newblock \emph{arXiv preprint arXiv:2303.08774}.

\bibitem[{Ouyang et~al.(2022)Ouyang, Wu, Jiang, Almeida, Wainwright, Mishkin, Zhang, Agarwal, Slama, Ray et~al.}]{instructgpt}
Long Ouyang, Jeffrey Wu, Xu~Jiang, Diogo Almeida, Carroll Wainwright, Pamela Mishkin, Chong Zhang, Sandhini Agarwal, Katarina Slama, Alex Ray, et~al. 2022.
\newblock Training language models to follow instructions with human feedback.
\newblock \emph{Advances in neural information processing systems}, 35:27730--27744.

\bibitem[{Paul et~al.(2021)Paul, Ganguli, and Dziugaite}]{el2n_grand}
Mansheej Paul, Surya Ganguli, and Gintare~Karolina Dziugaite. 2021.
\newblock Deep learning on a data diet: Finding important examples early in training.
\newblock \emph{Advances in neural information processing systems}, 34:20596--20607.

\bibitem[{Raffel et~al.(2020)Raffel, Shazeer, Roberts, Lee, Narang, Matena, Zhou, Li, and Liu}]{raffel2020exploring}
Colin Raffel, Noam Shazeer, Adam Roberts, Katherine Lee, Sharan Narang, Michael Matena, Yanqi Zhou, Wei Li, and Peter~J Liu. 2020.
\newblock Exploring the limits of transfer learning with a unified text-to-text transformer.
\newblock \emph{Journal of machine learning research}, 21(140):1--67.

\bibitem[{Sorscher et~al.(2022)Sorscher, Geirhos, Shekhar, Ganguli, and Morcos}]{sorscher2022beyond}
Ben Sorscher, Robert Geirhos, Shashank Shekhar, Surya Ganguli, and Ari Morcos. 2022.
\newblock Beyond neural scaling laws: beating power law scaling via data pruning.
\newblock \emph{Advances in Neural Information Processing Systems}, 35:19523--19536.

\bibitem[{Tan et~al.(2024)Tan, Wu, Du, Chen, Wang, Wang, and Qi}]{tan2024data}
Haoru Tan, Sitong Wu, Fei Du, Yukang Chen, Zhibin Wang, Fan Wang, and Xiaojuan Qi. 2024.
\newblock Data pruning via moving-one-sample-out.
\newblock \emph{Advances in Neural Information Processing Systems}, 36.

\bibitem[{Team(2024)}]{qwen2.5}
Qwen Team. 2024.
\newblock \href {https://qwenlm.github.io/blog/qwen2.5/} {Qwen2.5: A party of foundation models}.

\bibitem[{Tran et~al.(2024)Tran, Yang, Yao, and Yu}]{bioinstruct}
Hieu Tran, Zhichao Yang, Zonghai Yao, and Hong Yu. 2024.
\newblock Bioinstruct: instruction tuning of large language models for biomedical natural language processing.
\newblock \emph{Journal of the American Medical Informatics Association}, page ocae122.

\bibitem[{{\"U}st{\"u}n et~al.(2024){\"U}st{\"u}n, Aryabumi, Yong, Ko, D'souza, Onilude, Bhandari, Singh, Ooi, Kayid et~al.}]{ustun2024aya}
Ahmet {\"U}st{\"u}n, Viraat Aryabumi, Zheng-Xin Yong, Wei-Yin Ko, Daniel D'souza, Gbemileke Onilude, Neel Bhandari, Shivalika Singh, Hui-Lee Ooi, Amr Kayid, et~al. 2024.
\newblock Aya model: An instruction finetuned open-access multilingual language model.
\newblock \emph{arXiv preprint arXiv:2402.07827}.

\bibitem[{Wang et~al.(2025{\natexlab{a}})Wang, Yang, Wang, Li, Zhang, and Yan}]{SDC}
Shaobo Wang, Yantai Yang, Qilong Wang, Kaixin Li, Linfeng Zhang, and Junchi Yan. 2025{\natexlab{a}}.
\newblock Not all samples should be utilized equally: Towards understanding and improving dataset distillation.
\newblock In \emph{Synthetic Data for Computer Vision Workshop @ CVPR 2025}.

\bibitem[{Wang et~al.(2025{\natexlab{b}})Wang, Yang, Zhang, Sun, Li, Hu, and Zhang}]{DRUPI}
Shaobo Wang, Yantai Yang, Shuaiyu Zhang, Chenghao Sun, Weiya Li, Xuming Hu, and Linfeng Zhang. 2025{\natexlab{b}}.
\newblock {DRUPI}: Dataset reduction using privileged information.
\newblock In \emph{The Future of Machine Learning Data Practices and Repositories at ICLR 2025}.

\bibitem[{Wang et~al.(2025{\natexlab{c}})Wang, Yang, Liu, Sun, Hu, He, and Zhang}]{NCFM}
Shaobo Wang, Yicun Yang, Zhiyuan Liu, Chenghao Sun, Xuming Hu, Conghui He, and Linfeng Zhang. 2025{\natexlab{c}}.
\newblock Dataset distillation with neural characteristic function: A minmax perspective.
\newblock In \emph{Proceedings of the IEEE conference on computer vision and pattern recognition}.

\bibitem[{Wang et~al.(2020)Wang, Zhu, Torralba, and Efros}]{DD}
Tongzhou Wang, Jun-Yan Zhu, Antonio Torralba, and Alexei~A. Efros. 2020.
\newblock \href {https://arxiv.org/abs/1811.10959} {Dataset distillation}.
\newblock \emph{Preprint}, arXiv:1811.10959.

\bibitem[{Wang et~al.(2022)Wang, Kordi, Mishra, Liu, Smith, Khashabi, and Hajishirzi}]{wang2022self}
Yizhong Wang, Yeganeh Kordi, Swaroop Mishra, Alisa Liu, Noah~A Smith, Daniel Khashabi, and Hannaneh Hajishirzi. 2022.
\newblock Self-instruct: Aligning language models with self-generated instructions.
\newblock \emph{arXiv preprint arXiv:2212.10560}.

\bibitem[{Wei et~al.(2021)Wei, Bosma, Zhao, Guu, Yu, Lester, Du, Dai, and Le}]{wei2021finetuned}
Jason Wei, Maarten Bosma, Vincent~Y Zhao, Kelvin Guu, Adams~Wei Yu, Brian Lester, Nan Du, Andrew~M Dai, and Quoc~V Le. 2021.
\newblock Finetuned language models are zero-shot learners.
\newblock \emph{arXiv preprint arXiv:2109.01652}.

\bibitem[{Wendler et~al.(2024)Wendler, Veselovsky, Monea, and West}]{wendler2024llamas}
Chris Wendler, Veniamin Veselovsky, Giovanni Monea, and Robert West. 2024.
\newblock Do llamas work in english? on the latent language of multilingual transformers.
\newblock \emph{arXiv preprint arXiv:2402.10588}.

\bibitem[{Xia et~al.(2024)Xia, Malladi, Gururangan, Arora, and Chen}]{LESS}
Mengzhou Xia, Sadhika Malladi, Suchin Gururangan, Sanjeev Arora, and Danqi Chen. 2024.
\newblock Less: Selecting influential data for targeted instruction tuning.
\newblock \emph{arXiv preprint arXiv:2402.04333}.

\bibitem[{Xu et~al.(2024)Xu, Jiang, Niu, Deng, Poovendran, Choi, and Lin}]{xu2024magpie}
Zhangchen Xu, Fengqing Jiang, Luyao Niu, Yuntian Deng, Radha Poovendran, Yejin Choi, and Bill~Yuchen Lin. 2024.
\newblock Magpie: Alignment data synthesis from scratch by prompting aligned llms with nothing.
\newblock \emph{arXiv preprint arXiv:2406.08464}.

\bibitem[{Yang et~al.(2024{\natexlab{a}})Yang, Yang, Hui, Zheng, Yu, Zhou, Li, Li, Liu, Huang, Dong, Wei, Lin, Tang, Wang, Yang, Tu, Zhang, Ma, Xu, Zhou, Bai, He, Lin, Dang, Lu, Chen, Yang, Li, Xue, Ni, Zhang, Wang, Peng, Men, Gao, Lin, Wang, Bai, Tan, Zhu, Li, Liu, Ge, Deng, Zhou, Ren, Zhang, Wei, Ren, Fan, Yao, Zhang, Wan, Chu, Liu, Cui, Zhang, and Fan}]{qwen2}
An~Yang, Baosong Yang, Binyuan Hui, Bo~Zheng, Bowen Yu, Chang Zhou, Chengpeng Li, Chengyuan Li, Dayiheng Liu, Fei Huang, Guanting Dong, Haoran Wei, Huan Lin, Jialong Tang, Jialin Wang, Jian Yang, Jianhong Tu, Jianwei Zhang, Jianxin Ma, Jin Xu, Jingren Zhou, Jinze Bai, Jinzheng He, Junyang Lin, Kai Dang, Keming Lu, Keqin Chen, Kexin Yang, Mei Li, Mingfeng Xue, Na~Ni, Pei Zhang, Peng Wang, Ru~Peng, Rui Men, Ruize Gao, Runji Lin, Shijie Wang, Shuai Bai, Sinan Tan, Tianhang Zhu, Tianhao Li, Tianyu Liu, Wenbin Ge, Xiaodong Deng, Xiaohuan Zhou, Xingzhang Ren, Xinyu Zhang, Xipin Wei, Xuancheng Ren, Yang Fan, Yang Yao, Yichang Zhang, Yu~Wan, Yunfei Chu, Yuqiong Liu, Zeyu Cui, Zhenru Zhang, and Zhihao Fan. 2024{\natexlab{a}}.
\newblock Qwen2 technical report.
\newblock \emph{arXiv preprint arXiv:2407.10671}.

\bibitem[{Yang et~al.(2024{\natexlab{b}})Yang, Ye, Ouyang, Zhou, and Shen}]{yang2024clippoweredframeworkrobustgeneralizable}
Suorong Yang, Peng Ye, Wanli Ouyang, Dongzhan Zhou, and Furao Shen. 2024{\natexlab{b}}.
\newblock \href {https://arxiv.org/abs/2410.11215} {A clip-powered framework for robust and generalizable data selection}.
\newblock \emph{Preprint}, arXiv:2410.11215.

\bibitem[{Yang et~al.(2024{\natexlab{c}})Yang, Mishra, Chiang, and Mirzasoleiman}]{S2L}
Yu~Yang, Siddhartha Mishra, Jeffrey~N Chiang, and Baharan Mirzasoleiman. 2024{\natexlab{c}}.
\newblock Smalltolarge (s2l): Scalable data selection for fine-tuning large language models by summarizing training trajectories of small models.
\newblock \emph{arXiv preprint arXiv:2403.07384}.

\bibitem[{Yu et~al.(2024)Yu, Chen, Ahmadian, and Fadaee}]{yu2024diversify}
Simon Yu, Liangyu Chen, Sara Ahmadian, and Marzieh Fadaee. 2024.
\newblock Diversify and conquer: Diversity-centric data selection with iterative refinement.
\newblock \emph{arXiv preprint arXiv:2409.11378}.

\bibitem[{Yu et~al.(2025)Yu, Zhang, Zhang, Liang, Zhang, Zhang, Khademi, Awadalla, Wang, Yang, and Wei}]{yu2025chainofreasoningunifiedmathematicalreasoning}
Yiyao Yu, Yuxiang Zhang, Dongdong Zhang, Xiao Liang, Hengyuan Zhang, Xingxing Zhang, Mahmoud Khademi, Hany Awadalla, Junjie Wang, Yujiu Yang, and Furu Wei. 2025.
\newblock \href {https://arxiv.org/abs/2501.11110} {Chain-of-reasoning: Towards unified mathematical reasoning in large language models via a multi-paradigm perspective}.
\newblock \emph{Preprint}, arXiv:2501.11110.

\bibitem[{Zhang et~al.(2023)Zhang, Dong, Li, Zhang, Sun, Wang, Li, Hu, Zhang, Wu et~al.}]{ITsurvey}
Shengyu Zhang, Linfeng Dong, Xiaoya Li, Sen Zhang, Xiaofei Sun, Shuhe Wang, Jiwei Li, Runyi Hu, Tianwei Zhang, Fei Wu, et~al. 2023.
\newblock Instruction tuning for large language models: A survey.
\newblock \emph{arXiv preprint arXiv:2308.10792}.

\bibitem[{Zhang et~al.(2024)Zhang, Zhai, Ma, Shen, Li, Jiang, and Liu}]{staff}
Xiaoyu Zhang, Juan Zhai, Shiqing Ma, Chao Shen, Tianlin Li, Weipeng Jiang, and Yang Liu. 2024.
\newblock Speculative coreset selection for task-specific fine-tuning.
\newblock \emph{arXiv preprint arXiv:2410.01296}.

\bibitem[{Zhao et~al.(2021)Zhao, Mopuri, and Bilen}]{DC}
Bo~Zhao, Konda~Reddy Mopuri, and Hakan Bilen. 2021.
\newblock Dataset condensation with gradient matching.
\newblock In \emph{Ninth International Conference on Learning Representations 2021}.

\bibitem[{Zheng et~al.(2023)Zheng, Liu, Lai, and Prakash}]{ccs}
Haizhong Zheng, Rui Liu, Fan Lai, and Atul Prakash. 2023.
\newblock Coverage-centric coreset selection for high pruning rates.
\newblock In \emph{The Eleventh International Conference on Learning Representations}.

\bibitem[{Zheng et~al.(2024)Zheng, Zhang, Zhang, Ye, Luo, Feng, and Ma}]{llamafactory}
Yaowei Zheng, Richong Zhang, Junhao Zhang, Yanhan Ye, Zheyan Luo, Zhangchi Feng, and Yongqiang Ma. 2024.
\newblock \href {https://arxiv.org/abs/2403.13372} {Llamafactory: Unified efficient fine-tuning of 100+ language models}.
\newblock \emph{Preprint}, arXiv:2403.13372.

\bibitem[{Zhou et~al.(2024)Zhou, Frank, and McCoy}]{zhou2024context}
Zhenghao Zhou, Robert Frank, and R~Thomas McCoy. 2024.
\newblock Is in-context learning a type of gradient-based learning? evidence from the inverse frequency effect in structural priming.
\newblock \emph{arXiv preprint arXiv:2406.18501}.

\end{thebibliography}

\clearpage

\appendix

\section{More Experimental Details}\label{sec:more_exp}

\subsection{Datasets}\label{sec:dataset}

We use three real-world datasets for corresponding downstream tasks, namely, the BioInstruct dataset~\citep{bioinstruct} for biomedical question answering, the DialogSum dataset~\citep{dialogsum} for dialogue summarization, and GSM8K~\citep{gsm8k} for mathematical reasoning. The BioInstruct dataset contains 25,005 instruction-input-output triplets spanning diverse biomedical scenarios, including clinical decision-making, biomedical question answering, and diagnostic interpretation. We split the dataset into training and testing sets with a 9:1 ratio. The DialogSum dataset is a comprehensive collection of 13,460 dialogues curated from open dialogue repositories, addressing a variety of everyday scenarios. Each dialogue is paired with a carefully constructed reference summary, enabling robust evaluation of summarization models.  The GSM8K dataset offers 8.5K linguistically diverse, high-quality grade school math word problems requiring multi-step arithmetic reasoning. This dataset is designed to facilitate research on multi-step reasoning in question answering tasks for fundamental mathematical problem solving.

\subsection{In-Context Learning Prompt Template}\label{sec:ICL_prompt}

\begin{figure*}[tb!]
    \vspace{-20pt}
    \centering
    \includegraphics[width=0.85\linewidth]{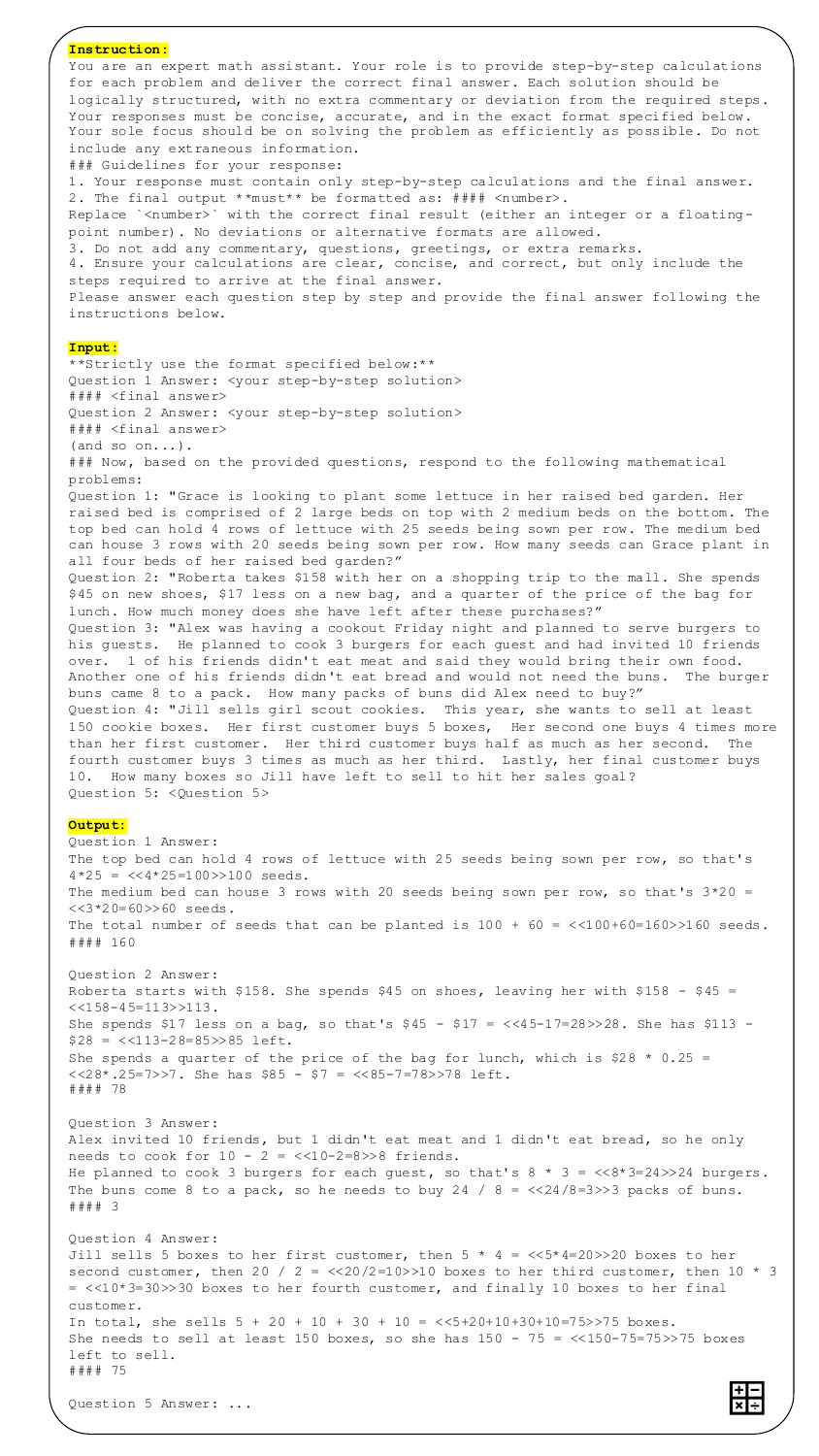}
    \vspace{-15pt}
    \caption{ICL prompt of GSM8k dataset.}
    \label{fig:icl_gsm8k}
\end{figure*}

\begin{figure*}[tb!]
    \centering
    \vspace{-20pt}
    \includegraphics[width=0.85\linewidth]{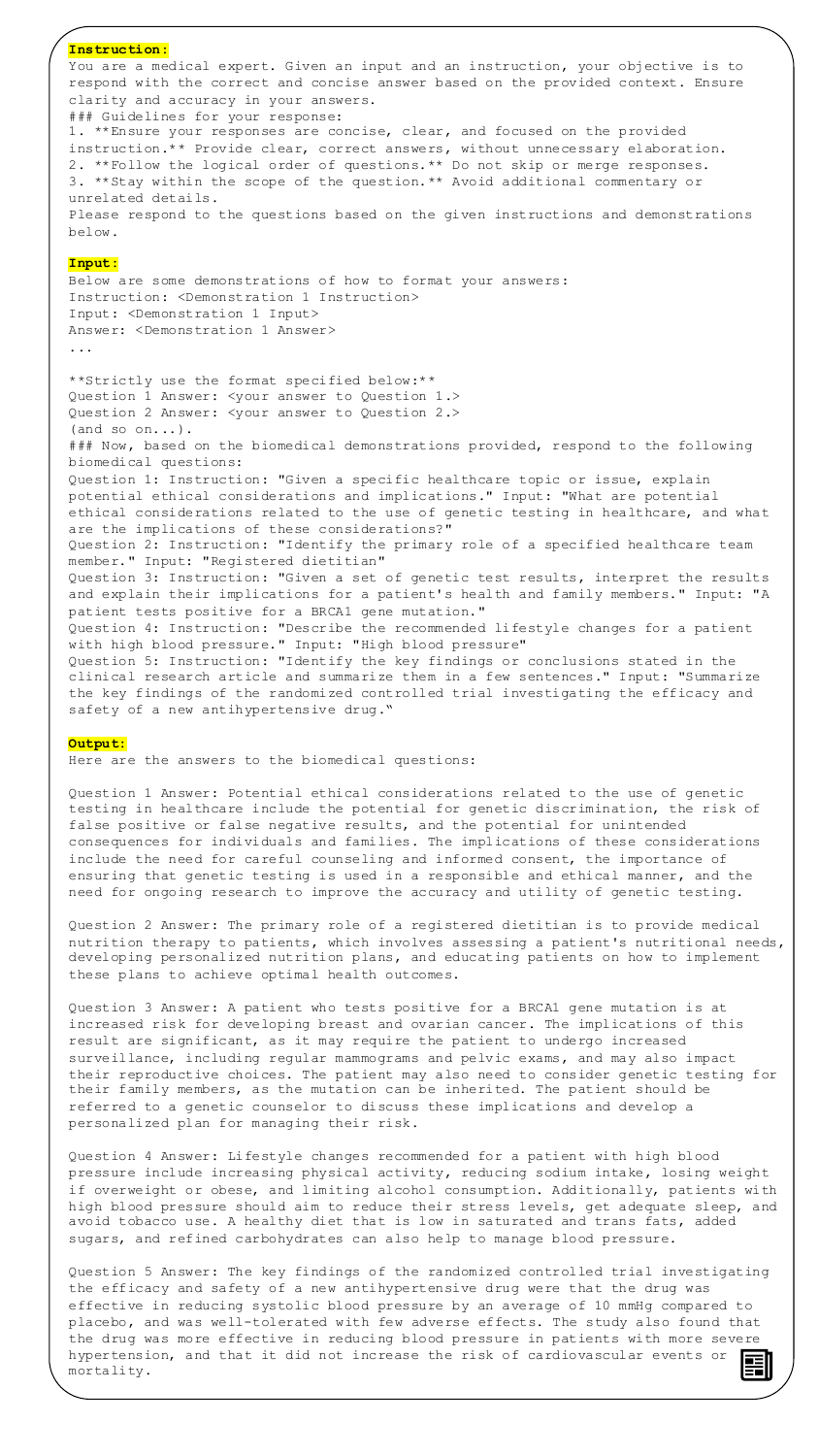}
    \vspace{-15pt}
    \caption{ICL prompt of DialogSum dataset.}
    \label{fig:icl_dialog}
\end{figure*}

\begin{figure*}[tb!]
    \centering
    \vspace{-20pt}
    \includegraphics[width=0.85\linewidth]{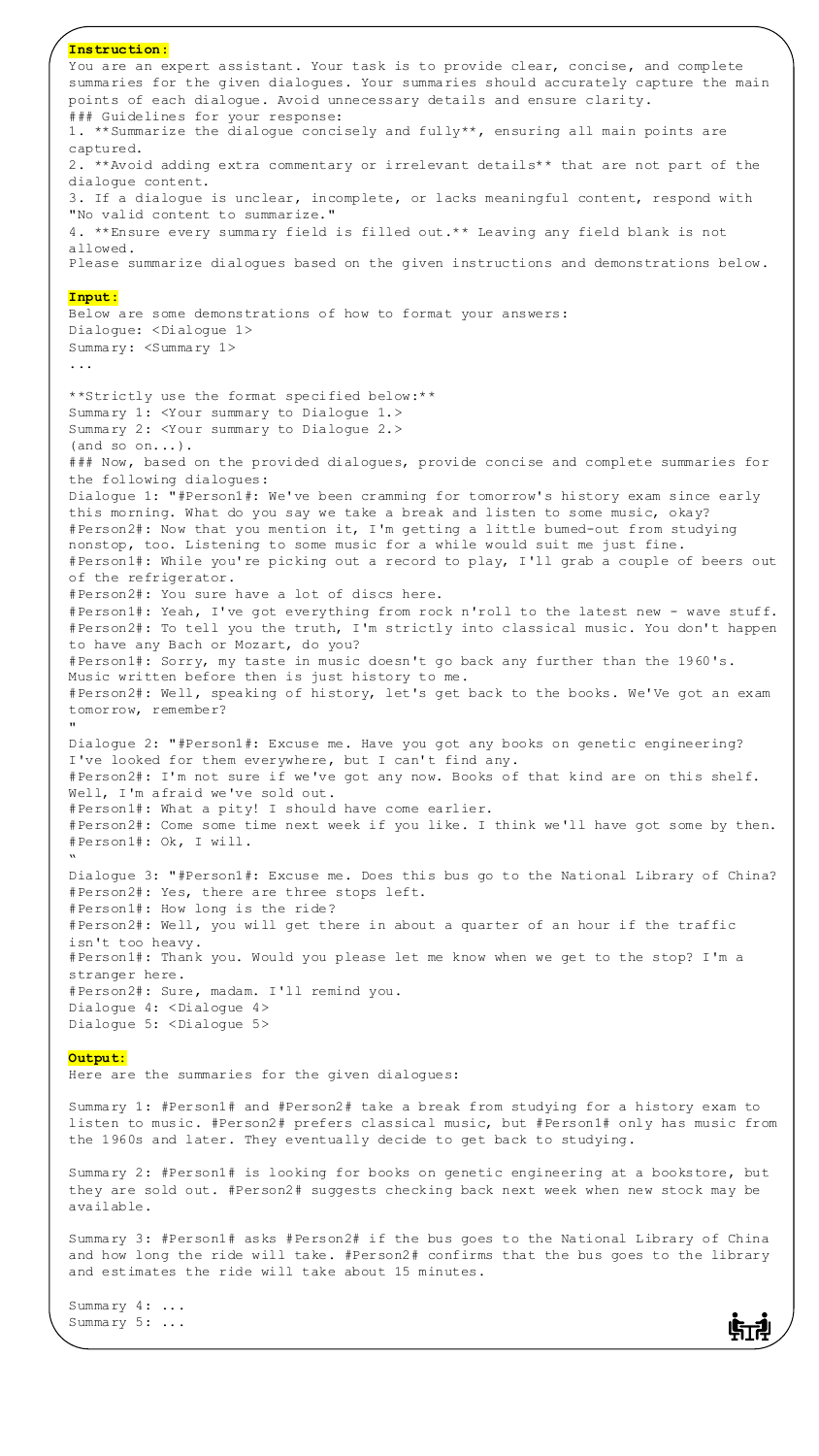}
    \vspace{-15pt}
    \caption{ICL prompt of BioInstruct dataset.}
    \label{fig:icl_bio}
\end{figure*}

We formalize the design of In-Context Learning templates for three datasets: BioInstruct, DialogSum, and GSM8K. Each template is structured to provide explicit guidance through task-specific instructions, demonstrations, and output constraints. 

\noindent \textbf{Template Design.} Our framework employs a unified template structure comprising three core components: (i) \textbf{Task Instruction.} A declarative statement that explicitly defines the task objective (\emph{e.g.,} "Generate a concise summary of the dialogue") alongside specific constraints (\emph{e.g.,} brevity, precision). (ii) \textbf{In-Context Demonstrations.} A set of input-output pairs that exemplify valid task execution, ensuring alignment with the task's requirements and constraints. (iii) \textbf{Output Constraints.} Rigorous formatting rules that enforce compliance with task-specific syntax, such as step-by-step derivations for GSM8K or speaker-aware summarization for DialogSum. This modular architecture ensures consistency and adaptability across diverse tasks while maintaining strict adherence to domain-specific guidelines.

\noindent \textbf{Task-Specific Prompt Design.} We present a systematic approach to designing In-Context Learning prompts tailored to diverse datasets. For BioInstruct, we prioritize biomedical accuracy by implementing hierarchical guidelines that emphasize logical ordering and scope limitation, while ensuring responses are strictly evidence-based and devoid of speculative content. In DialogSum, we enforce structured dialogue processing through explicit speaker tagging (\emph{e.g.,} \texttt{\#Person1\#}), discourse compression techniques, and the generation of neutral third-person summaries. For GSM8K, we mandate a rigorous format that includes intermediate arithmetic steps and requires final answers to be boxed (\emph{e.g.,} \texttt{\#\#\#\# <number>}), explicitly prohibiting textual explanations to maintain precision and clarity. This tailored approach ensures that each dataset's unique characteristics are effectively leveraged to optimize In-Context Learning performance. Structural schematics and formatting rules are visualized in Figure~\ref{fig:icl_gsm8k}, ~\ref{fig:icl_dialog} and ~\ref{fig:icl_bio}. 

\clearpage
\subsection{Synthetic Data Generation}\label{sec:synthetic}

\begin{figure*}[tb!]
    \centering
    \includegraphics[width=0.99\linewidth]{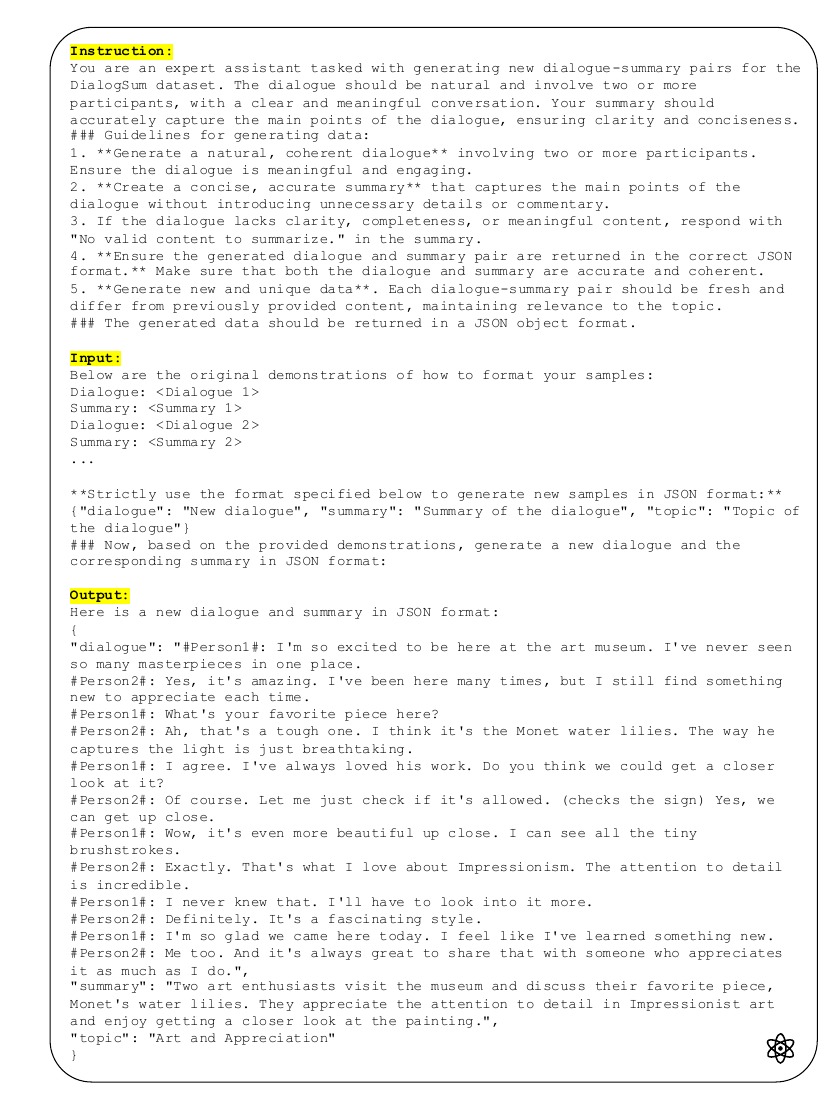}
    \vspace{-15pt}
    \caption{Data synthesis prompt of DialogSum dataset.}
    \label{fig:icl_bio_syn}
\end{figure*}

To evaluate the robustness of \mymethod{} on synthetic data, we constructed a synthetic variant of the DialogSum dataset using the Llama-3-8B-Instruct model. The generation pipeline comprised three key stages: \textit{prompt design}, \textit{generation process}, and \textit{quality control}.

\noindent \textbf{Prompt Design.} For every 5 demonstration samples from the original DialogSum dataset, we crafted structured prompts to generate 1 synthetic dialogue-summary pair. These prompts enforced strict formatting rules, including speaker tagging and topic alignment, alongside explicit constraints to ensure the synthetic data closely mirrored the style and structure of the original dataset.

\noindent \textbf{Generation Process.} The model was instructed to produce dialogues with multiple turns, ensuring natural conversational flow and coherent narratives. Each output adhered to the JSON schema, with topics covering diverse everyday scenarios such as travel, shopping, and education.

\noindent \textbf{Quality Control.} A two-step filtering process was employed to ensure data quality. First, automated checks removed duplicates and syntax-invalid entries using regex-based validation. Second, three annotators independently reviewed the samples, flagging those with coherence issues, factual inconsistencies, or formatting violations. Disagreements were resolved through majority voting, and only samples free from hallucination or structural deviations were retained.

The final synthetic dataset maintained the original DialogSum distribution in dialogue length and topic diversity. Full prompt template is illustrated in Figure~\ref{fig:icl_bio_syn}.

\subsection{Hyper-Parameters of Different Methods}\label{sec:hyper_param}

\noindent \textbf{Data Selection.}  
In our experiments, \mymethod{} was configured with task-specific batch sizes: 15 for the BioInstruct dataset, 5 for the DialogSum dataset, and 15 for the GSM8K dataset. By default, $n_d$ (number of demonstrations) was set to 10, and $n_q$ (number of queries) was set to 5. The generation temperature was fixed at 0 during the In-Context Learning process to ensure deterministic outputs. To ensure fairness, the batch size for all baseline methods was uniformly set to 16 across all tasks.

\noindent \textbf{Model Fine-tuning.}  
All models, whether trained on pruned or full datasets, shared identical hyperparameters. We employed parameter-efficient fine-tuning via LoRA~\cite{lora} for all tasks. The fine-tuning framework, based on \cite{llamafactory}\footnote{\href{https://github.com/hiyouga/LLaMA-Factory}{https://github.com/hiyouga/LLaMA-Factory}}, utilized a learning rate scheduler with linear warm-up and cosine decay. A consistent batch size of 8 was applied across all experiments. For Llama-3-8B-Instruct, the learning rate was fixed at $1 \times 10^{-4}$ for all datasets. For Qwen-2.5-7B-Instruct and Mistral-Nemo-Instruct-2407, the learning rate was uniformly set to $1 \times 10^{-5}$. All models were fine-tuned for 5 epochs, with each experiment repeated three times to ensure robustness and statistical reliability.

\subsection{Pseudo Code of \mymethod{}}\label{sec:pseudo_code}

The detailed pseudo code of \mymethod{} is demonstrated in Algorithm~\ref{alg:method}.

\begin{algorithm*}[tb!]
\caption{\mymethod{} for Data Selection}
\textbf{Input:} \\
$\mathcal{M}_p$ -- Pre-trained language model \\
$\mathcal{D}$ -- Full task dataset \\
$n_d$ -- Number of samples for demonstration at a time \\
$n_q$ -- Number of samples for query at a time \\
$I$ -- Instruction for ICL generation \\
$f$ -- Performance metric \\
$\rho$ -- Budget ratio \\
\textbf{Output:} $\mathcal{D'}$ -- Selected coreset ($\mathcal{D'} \subseteq \mathcal{D}$, $|\mathcal{D'}| < |\mathcal{D}|$)
\begin{algorithmic}[1]
\Procedure{DataWhisperer}{$\mathcal{M}_p, \mathcal{D}, n_d, n_q, I, f, \rho$}
    \State $\mathcal{S} \gets \mathbf{0}$ \Comment{Initialize the score set}
    \While {Not all samples in $\mathcal{D}$ have been processed}
        \State $\mathcal{D}_d \gets \text{Randomly select } n_d \text{ samples from } \mathcal{D}$ 
        \State $\mathcal{D}_q \gets \text{Randomly select } n_q \text{ samples from } \mathcal{D}$ 
        \State $C \gets \{I, \mathcal{D}_d\}$ \Comment{Form input context}
        \State $\hat{y}_q^{(1)}, \dots, \hat{y}_q^{(n_q)} \gets \mathcal{M}_p(C, \mathcal{D}_q)$ \Comment{Generate query predictions via ICL}
        \State $s \gets \frac{1}{n_q}\sum_{j=1}^{n_q} f(\hat{y}_q^{(j)}, y_q^{(j)})$
        \For{$i = 1 \to n_d$}
            \State $\text{Let } h \text{ denote each head in the attention mechanism}$
            \State $w_i \gets \sum_h \mathbf{1}^\top A_{(x_d^{(i)}, y_d^{(i)})}^{(h)} \mathbf{1}$ \Comment{Compute weights}
        \EndFor
        \State $w^{*} \gets \text{Normalize}(w) $
        \State $s^{*} \gets s \odot w^{*}$ \Comment{Element-wise multiplication with attention weights}
        \State $\mathcal{S}[\mathcal{D}_d] \gets \mathcal{S}[\mathcal{D}_d] + s^{*}$ \Comment{Accumulate scores for corresponding indices in $\mathcal{D}_d$}
    \EndWhile
    \State $k \gets \lfloor \rho \cdot |\mathcal{D}| \rfloor$ 
    \State $\mathcal{D}' \gets \text{Top-k}(\mathcal{D};\mathcal{S})$ \Comment{Select top $m$ samples based on scores}
    \State \textbf{return} $\mathcal{D'}$ 
\EndProcedure
\end{algorithmic}
\label{alg:method}
\end{algorithm*}

\begin{table*}[tb!]
\centering
\vspace{-10pt}
\caption{Additional results on the Selection-to-Tuning ratio and performance scores for the Qwen and Mistral models are presented. The symbols \textcolor{PineGreen}{$\uparrow$} and \textcolor{OrangeRed}{$\downarrow$} indicate improvements and degradations compared to random selection, respectively. ``Speedup'' refers to the acceleration achieved by \mymethod{} (w2s) over Nuggets. ``w2s'' denotes using the weaker model within the same family for data selection across each LLM.} 
\vspace{-10pt}
\label{tab:str_appendix}
\resizebox{.99\textwidth}{!}{
\begin{tabular}{c|c|cccc|cccc|cccc}
\toprule
\multirow{3}{*}{\textbf{Model}} & \multirow{3}{*}{\textbf{Method}} & \multicolumn{4}{c|}{\textbf{GSM8K}}& \multicolumn{4}{c|}{\textbf{DialogSum}}& \multicolumn{4}{c}{\textbf{BioInstruct}}\\ 
 & &\multicolumn{3}{c}{\textbf{STR}}& \textbf{Performance}& \multicolumn{3}{c}{\textbf{STR}}& \textbf{Performance}& \multicolumn{3}{c}{\textbf{STR}}&\textbf{Performance}\\ 
 &&1\% & 5\% & 10\%     &Avg.  Score& 1\% & 5\% & 10\%     &Avg.  Score& 1\% & 5\% & 10\%     &Avg.  Score\\ \midrule
\parbox[t]{2mm}{\multirow{8}{*}{\rotatebox[origin=c]{90}{Qwen-2.5-7B-Instruct}}}
 & Random  & - & - & -  & 60.29 & - & -& - &  34.66 &- &- &- & 35.19\\
 & GraNd  & 1.06 & 1.06 & 1.07  & 62.61{\textcolor{PineGreen}{\footnotesize $\uparrow$2.32}} & 1.09 & 1.11 & 1.13  & 36.29{\textcolor{PineGreen}{\footnotesize $\uparrow$1.63}} & 1.07 & 1.08 & 1.34  & 32.62{\textcolor{OrangeRed}{\footnotesize $\downarrow$2.57}} \\
 & EL2N  &  1.06 & 1.06 & 1.07  & 60.72{\textcolor{PineGreen}{\footnotesize $\uparrow$0.43}} & 1.10 & 1.13 & 1.15  & 36.97{\textcolor{PineGreen}{\footnotesize $\uparrow$2.31}} & 1.08 & 1.08 & 1.32  & 34.64{\textcolor{OrangeRed}{\footnotesize $\downarrow$0.55}} \\
 & CCS  & 1.04 & 1.04 & 1.05  & 61.34{\textcolor{PineGreen}{\footnotesize $\uparrow$1.05}} & 1.01 & 1.04 & 1.06  & 35.30{\textcolor{PineGreen}{\footnotesize $\uparrow$0.64}} & 1.02 & 1.03 & 1.25  & 31.87{\textcolor{OrangeRed}{\footnotesize $\downarrow$3.32}} \\
 & Nuggets  & 1.10 & 1.11 & 1.11  & 62.74{\textcolor{PineGreen}{\footnotesize $\uparrow$2.45}} & 3.66 & 3.69 & 3.71  & 36.63{\textcolor{PineGreen}{\footnotesize $\uparrow$1.97}} & 2.19 & 2.20 & 2.41  & 34.95{\textcolor{OrangeRed}{\footnotesize $\downarrow$0.24}} \\ 
 & STAFF & 1.05 & 1.05 & 1.06  & 62.74{\textcolor{PineGreen}{\footnotesize $\uparrow$2.45}} & 1.06 & 1.08 & 1.10  & 37.18{\textcolor{PineGreen}{\footnotesize $\uparrow$2.52}} & 1.06 & 1.06 & 1.33  & 32.97{\textcolor{OrangeRed}{\footnotesize $\downarrow$2.22}} \\ 
 & \mymethod{} & \textbf{0.08} & \textbf{0.08} & \textbf{0.08}  & \textbf{64.21}{\textcolor{PineGreen}{\footnotesize $\uparrow$3.92}} & \textbf{0.16} & \textbf{0.19} & \textbf{0.29}  & \textbf{38.97}{\textcolor{PineGreen}{\footnotesize $\uparrow$4.31}} & \textbf{0.15} & \textbf{0.27} & \textbf{0.46}  & \textbf{37.12}{\textcolor{PineGreen}{\footnotesize $\uparrow$1.93}} \\ 
 & \mymethod{} (w2s) & \textbf{0.05} & \textbf{0.05} & \textbf{0.06}  & \textbf{63.67}{\textcolor{PineGreen}{\footnotesize $\uparrow$3.38}} & \textbf{0.10} & \textbf{0.13} & \textbf{0.23}  & \textbf{38.37}{\textcolor{PineGreen}{\footnotesize $\uparrow$3.71}} & \textbf{3.71} & \textbf{0.20} & \textbf{0.37}  & \textbf{35.98}{\textcolor{PineGreen}{\footnotesize $\uparrow$0.79}} \\ 
 & \cellcolor{mycolor}Speedup & \cellcolor{mycolor}\textbf{21.52$\times$} & \cellcolor{mycolor}\textbf{21.12$\times$} & \cellcolor{mycolor}\textbf{19.29$\times$}  & \cellcolor{mycolor}- & \cellcolor{mycolor}\textbf{35.67$\times$} & \cellcolor{mycolor}\textbf{27.48$\times$} & \cellcolor{mycolor}\textbf{16.12$\times$}  & \cellcolor{mycolor}- & \cellcolor{mycolor}\textbf{25.12$\times$} & \cellcolor{mycolor}\textbf{10.94$\times$} & \cellcolor{mycolor}\textbf{6.48$\times$}  & \cellcolor{mycolor}- \\ \midrule
\parbox[t]{2mm}{\multirow{8}{*}{\rotatebox[origin=c]{90}{Mistral-Nemo-2407-12B}}}
 & Random  & - & - & -  & 50.36 & - & -& - &  32.61 &- &- &- & 25.00\\
 & GraNd  & 1.08 & 1.09 & 1.24  & 51.65{\textcolor{PineGreen}{\footnotesize $\uparrow$1.29}} & 1.05 & 1.07 & 1.09  & 32.72{\textcolor{PineGreen}{\footnotesize $\uparrow$0.11}} & 1.13 & 1.23 & 1.40  & 24.35{\textcolor{OrangeRed}{\footnotesize $\downarrow$0.65}} \\
 & EL2N  &  1.11 & 1.13 & 1.36  & 51.74{\textcolor{PineGreen}{\footnotesize $\uparrow$1.38}} & 1.06 & 1.09 & 1.11  & 30.75{\textcolor{OrangeRed}{\footnotesize $\downarrow$1.86}} & 1.14 & 1.23 & 1.39  & 23.76{\textcolor{OrangeRed}{\footnotesize $\downarrow$1.24}} \\
 & CCS  & 1.04 & 1.01 & 1.22  & 50.71{\textcolor{PineGreen}{\footnotesize $\uparrow$0.35}} & 1.01 & 1.04 & 1.05  & 31.75{\textcolor{OrangeRed}{\footnotesize $\downarrow$0.86}} & 1.02 & 1.11 & 1.25  & 26.37{\textcolor{PineGreen}{\footnotesize $\uparrow$1.37}} \\
 & Nuggets  & 2.34 & 2.35 & 2.50  & 53.06{\textcolor{PineGreen}{\footnotesize $\uparrow$2.70}} & 2.22 & 2.24 & 2.26  & 32.53{\textcolor{OrangeRed}{\footnotesize $\downarrow$0.08}} & 5.15 & 5.23 & 5.36  & 28.12{\textcolor{PineGreen}{\footnotesize $\uparrow$3.12}} \\ 
 & STAFF  & 1.06 & 1.08 & 1.22  & 54.06{\textcolor{PineGreen}{\footnotesize $\uparrow$3.70}} & 1.04 & 1.06 & 1.07  & 32.71{\textcolor{PineGreen}{\footnotesize $\uparrow$0.10}}  & 1.10 & 1.20 & 1.39  & 25.03{\textcolor{PineGreen}{\footnotesize $\uparrow$0.03}} \\ 
 & \mymethod{} & \textbf{0.24} & \textbf{0.26} & \textbf{0.30}  & \textbf{57.62}{\textcolor{PineGreen}{\footnotesize $\uparrow$7.26}} & \textbf{0.08} & \textbf{0.12} & \textbf{0.25}  & \textbf{35.45}{\textcolor{PineGreen}{\footnotesize $\uparrow$2.84}} & \textbf{0.24} & \textbf{0.39} & \textbf{0.65}  & \textbf{29.96}{\textcolor{PineGreen}{\footnotesize $\uparrow$4.96}} \\ 
 & \mymethod{} (w2s) & \textbf{0.11} & \textbf{0.12} & \textbf{0.12}  & \textbf{57.21}{\textcolor{PineGreen}{\footnotesize $\uparrow$6.85}} & \textbf{0.07} & \textbf{0.11} & \textbf{0.24}  & \textbf{35.07}{\textcolor{PineGreen}{\footnotesize $\uparrow$2.46}} & \textbf{0.18} & \textbf{0.33} & \textbf{0.60}  & \textbf{29.66}{\textcolor{PineGreen}{\footnotesize $\uparrow$4.66}} \\ 
  & \cellcolor{mycolor}Speedup & \cellcolor{mycolor}\textbf{20.52$\times$} & \cellcolor{mycolor}\textbf{20.45$\times$} & \cellcolor{mycolor}\textbf{20.89$\times$}  & \cellcolor{mycolor}- & \cellcolor{mycolor}\textbf{31.75$\times$} & \cellcolor{mycolor}\textbf{20.69$\times$} & \cellcolor{mycolor}\textbf{9.53$\times$}  & \cellcolor{mycolor}- & \cellcolor{mycolor}\textbf{29.18$\times$} & \cellcolor{mycolor}\textbf{15.61$\times$} & \cellcolor{mycolor}\textbf{8.95$\times$}  & \cellcolor{mycolor}- \\ \bottomrule
\end{tabular}
}
\vspace{-10pt}
\end{table*}

\section{Extensive Evaluation Results of STR} \label{sec:STR}

As defined in Eq.~\eqref{eq:str}, the \textit{Selection-to-Tuning Ratio} (STR) quantifies the efficiency of data selection methods by comparing the computational cost of selection to the time saved during fine-tuning. A method $\tau$ is efficient if $\text{STR} < 1$, indicating that the time invested in selection is offset by faster fine-tuning. An STR below 1 ensures that the selection process positively contributes to overall pipeline efficiency, making it a practical and scalable choice for optimizing model training workflows. More evaluation results of STR are demonstrated in Table~\ref{tab:str_appendix}.

\end{document}